\documentclass[conference]{IEEEtran}
\IEEEoverridecommandlockouts
\usepackage{amsmath,amssymb,amsthm,booktabs,graphicx,xcolor,xspace,cite}
\usepackage{algorithm,algpseudocode}
\usepackage[hidelinks]{hyperref}
\newtheorem{proposition}{Proposition}

\newcommand{\lean}{\textsf{\textbf{LEAN}}\xspace}

\begin{document}

\title{LEAN: Loss-Engineered Modulation of the Model Weight-Sensitivity Distribution for Efficient, Error-Tolerant Neural Networks}

\author{\IEEEauthorblockN{Habibur Rahaman and Swarup Bhunia}
\IEEEauthorblockA{Electrical and Computer Engineering, University of Florida\\
Gainesville, Florida, USA\\
rahaman.habibur@ufl.edu, swarup@ece.ufl.edu}}
\maketitle

\begin{abstract}
The weights of a trained network do not have uniform sensitivity to model accuracy: the 1\% most sensitive weights carry about a fifth of the damage a bit flip can do. This distribution decides the bit-flip tolerance of a model, the fraction of weights a code must cover and the bits each weight needs; existing schemes work with the distribution training leaves. We show that we can systematically modulate or alter the weight sensitivity distribution of a trained model, thereby shifting weights from one sensitivity bin to another. The sensitivity of a stored weight is the product of its gradient and the step a flip takes in the storage format, and \lean adds a composite regularizer on both factors, a range term that bounds the flip step and a tail term that lowers the gradient of the most sensitive weights, with one coefficient $\lambda$ setting how far the distribution moves. Ten epochs with the term, as a fine-tune or as the last epochs of training, shrink the sensitive tail 5 to 13 times on nine vision models (ResNet-18, ResNet-50, VGG-16 on CIFAR-10, CIFAR-100, ImageNet) within 0.5 points of accuracy; real flips confirm that the reshaped weights do less damage, a plain gradient-norm penalty does not reproduce the effect, and a preliminary extension to three small language models shows the same tail shaping. With a BCH code over 1\% of the weights the model then survives 300 progressive bit-search flips in every training seed and 500 in two of three, where three of four published defenses with the same code fail by 300 and WCAT falls to 75\% at 500. The reshaped distribution also cuts the parity for a fixed guarantee 3.7 times and makes a sensitivity-ordered 8/6/4/3-bit format affordable, which stores ImageNet ResNet-50 at a 3.29-bit weight payload, 2.4 times smaller than INT8, above the accuracy of LSQ at 3 bits.
\end{abstract}


\begin{figure}[!t]
\centering\includegraphics[width=\columnwidth]{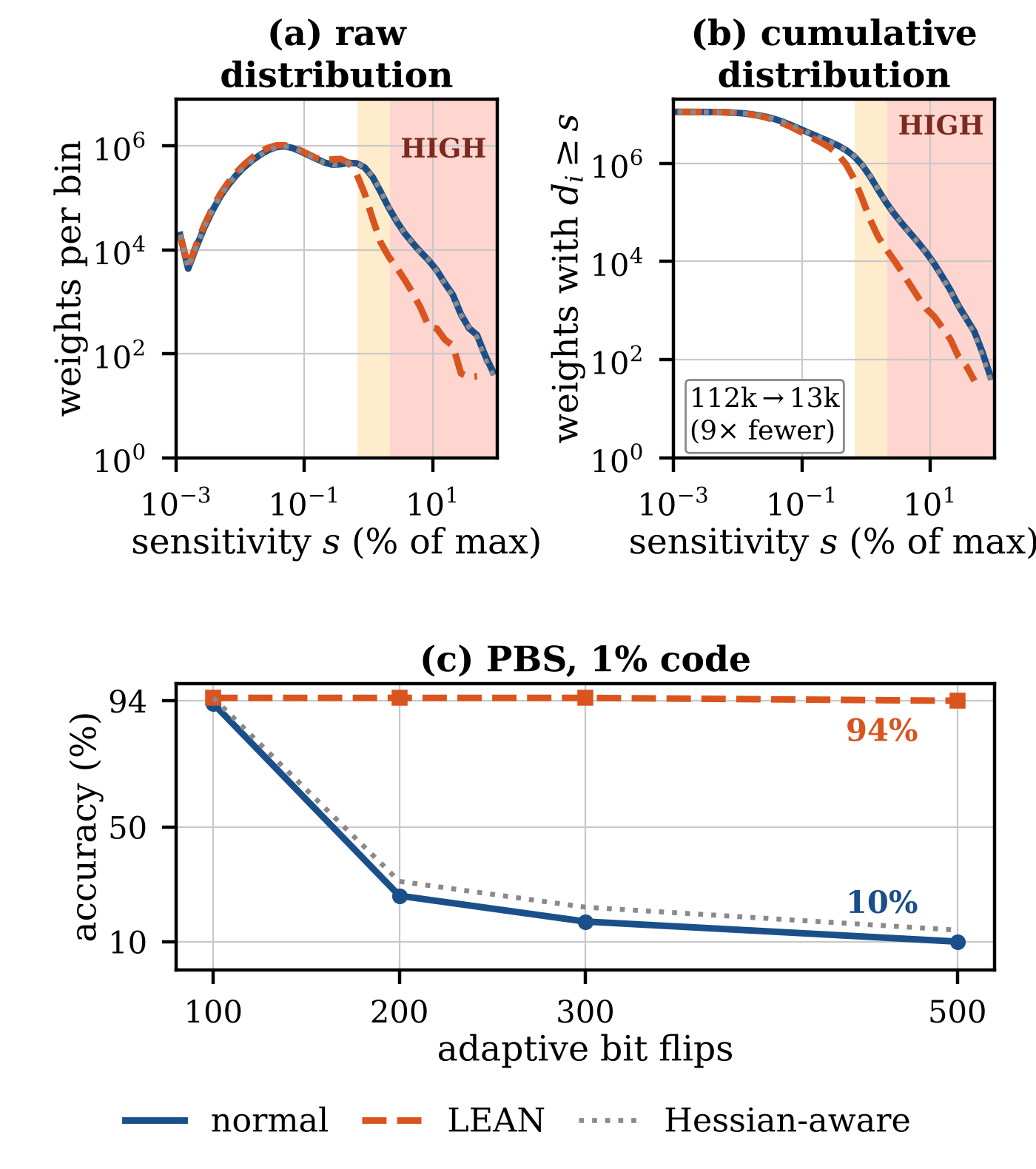}
\caption{Effect of \lean on ResNet-18, CIFAR-10. (a) The raw weight-sensitivity distribution: number of weights per sensitivity bin (40 log-spaced bins of the sign-flip damage $d_i$, as a percentage of the baseline maximum); the bulk of the weights sits between 0.01 and 1\% and the HIGH band is a thin tail, which \lean moves to the left. (b) The cumulative distribution, counted from the tail: the number of weights whose damage is at least a threshold $s$; one ten-epoch fine-tune leaves 9 times fewer weights above the baseline HIGH threshold at the same accuracy, while Hessian-aware training leaves the tail where normal training left it (1.01\% against 1.00\% above the threshold). (c) Accuracy under progressive bit search when the same 1\% BCH code protects each model: the normally trained and the Hessian-aware model fall to chance by 300 flips, the \lean model ($\lambda=40$) keeps 94\% after 500 in two of three training seeds and after 300 in all (Table~\ref{tab:def}).}
\label{fig:teaser}
\end{figure}
\begin{figure*}[t]
\centering\includegraphics[width=\textwidth]{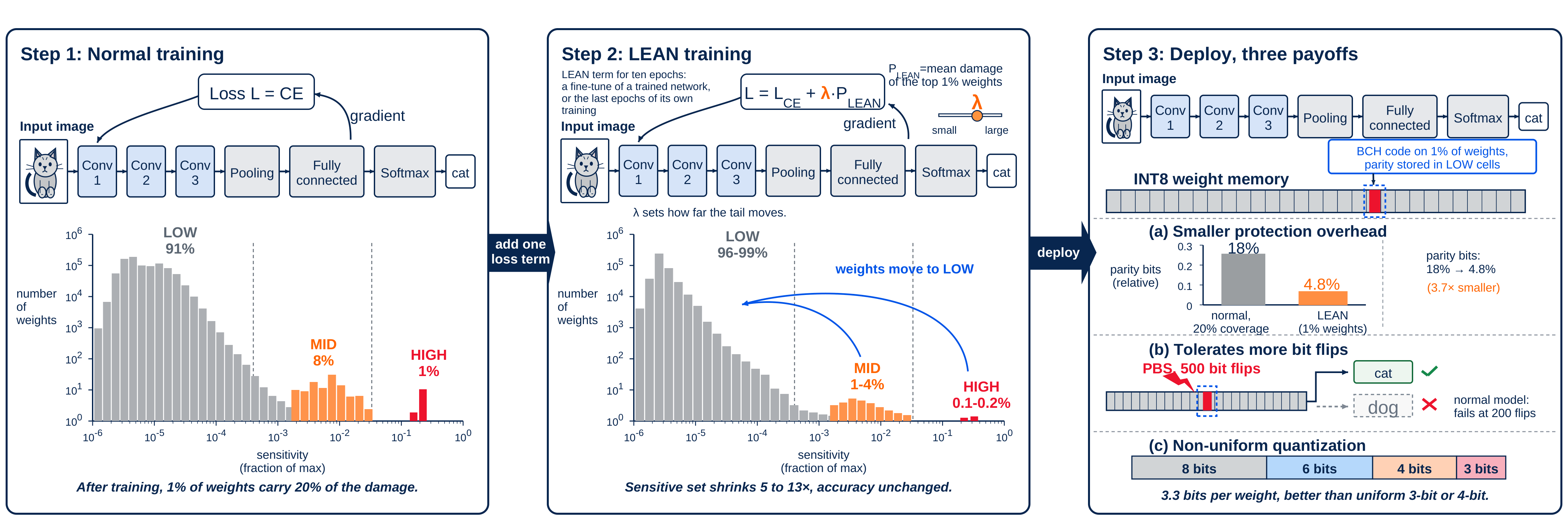}
\caption{Overview. Step 1: a normally trained network has a heavy-tailed weight-sensitivity distribution; the top 1\% of weights (HIGH) carry about 20\% of the first-order damage of sign-bit flips. Step 2: ten epochs with the extra term $\lambda P_{\mathrm{LEAN}}$, as a fine-tune or as the last epochs of training, move weights out of HIGH and MID into LOW; $\lambda$ sets how far, and accuracy changes by at most 0.5 points. Step 3: the deployed model needs 3.7 times less parity for the same guarantee, survives 300 to 500 progressive bit-search flips with a 1\% BCH code, and stores at 3.3 bits per weight in a sensitivity-ordered format.}
\label{fig:overview}
\end{figure*}

\section{Introduction}
Modern neural networks have tens of millions to billions of weights and they do not impact model functional behavior uniformly. If a single stored weight is corrupted, by a memory fault, a bit-flip attack or a coarse quantization step, the loss change depends on which weight it is, and it differs by a factor of 10,000 or more between the median weight and the most sensitive one. On ResNet-18 trained on CIFAR-10, the 1\% of weights with the largest first-order sensitivity account for 19.5\% of the total loss change under sign-bit flips, the next 8\% for most of the rest, and the remaining 91\% contribute little under this metric (Figure~\ref{fig:teaser}a,b). Hence, weight sensitivity is strongly non-uniform, and its distribution has a large spread between the highest and the lowest sensitivity.

The shape of that distribution influences the cost of the stored model in terms of error-tolerance and computation efficiency. It sets bit-flip tolerance, because Rowhammer \cite{kim2014rowhammer,yao2020deephammer}, progressive bit search \cite{rakin2019bfa}, its targeted variant \cite{rakin2021tbfa} and low-voltage SRAM errors \cite{stutz2021bit} all do their damage through the tail (the attack surface of AI hardware is surveyed in \cite{rahaman2026evolving}), and an adaptive attacker only needs to hit it. It sets the price of protection, because a selective code \cite{guan2019inplace} has to cover the tail and its parity grows with the tail's size, while a code over everything costs 12 to 20\% of memory. And it sets the price of storage, because a variable-precision format must keep the tail at full width and can be cheap only on the rest. A heavier tail means more weights to protect, more parity, more bits and a shorter survival; a lighter one means the opposite. Robust training \cite{he2020pc,chitsaz2023wcat,nazari2024far,pranto2025hessian} changes the distribution as a side effect, and selective protection and mixed precision live with whatever distribution they are given; none of them optimizes the size of the tail, measured in the format the weights are stored in.

This paper treats the sensitivity distribution as something to be chosen. \lean is a training-time procedure that reshapes it. The damage of a flip is the product of two factors, the gradient of the weight and the step the flip takes in the storage format, and \lean regularizes both: a range term bounds the step of every layer, and a tail term, the mean sensitivity of the top 1\% of weights differentiated through the gradient, lowers the gradient where it is largest. A coefficient $\lambda$ sets how far the tail is pushed (Figure~\ref{fig:overview}). The change is to the whole distribution, not to a handful of weights, and it costs essentially no accuracy. The paper includes two checks: real flips confirm that the reshaped weights do less damage, not only that their gradients read smaller, and an ablation separates the two terms. At INT8 the range term does most of the work; the tail term adds to it but cannot replace it, and alone it lowers the gradient without lowering the damage. Figure~\ref{fig:teaser} previews the effect on one model: the tail shrinks by an order of magnitude, and under the same 1\% code the model survives an attack that reduces the normally trained one to chance.

The contributions of this paper are the following.
\begin{itemize}\setlength{\itemsep}{1pt}
\item \textbf{A systematic way to modulate the weight-sensitivity distribution of a network during training.} One composite regularizer on the two factors of the damage, with a coefficient $\lambda$ that sets the amount, an ablation that attributes the effect to each term, a real-flip check that the effect is not gradient masking (Table~\ref{tab:ablate}), and a plain gradient-norm baseline that does not reproduce it. Measured on nine vision model-dataset configurations (ResNet-18, ResNet-50 and VGG-16 on CIFAR-10, CIFAR-100 and ImageNet) and three language models (Qwen2-0.5B, Llama-3.2-1B and Qwen2-1.5B): the sensitive tail shrinks 5 to 13 times on the vision models within 0.5 points of accuracy, and 1.7 to 5.3 times on the language models, after one fine-tune and monotonically in $\lambda$. It is applied as a ten-epoch fine-tune of a trained network or switched on in the last epochs of training; both routes reach the same floor.
\item \textbf{Consequences of the reshaped distribution for protection, precision and storage.} Bit-flip protection: a 1\% code that fails on the normal model by 200 flips holds through 300 in every run and 94\% through 500 in two of three seeds, and the parity for a fixed guarantee drops 3.7 times. Precision and storage: a sensitivity-ordered 8/6/4/3-bit format is 1.2 points above LSQ on ImageNet ResNet-50 at an exactly matched 4 bits per weight, and at 3.29 bits reaches 77.2\% against 74.4\% for LSQ at 3 bits, 2.4 times less weight payload than INT8 and an estimated 59\% less weight-fetch energy; most of that gain comes from ordering the bits by sensitivity, which the small tail makes affordable, and $\lambda$ adds the rest. With a small interleaved code the CIFAR low-bit models survive 500 flips as well. The evaluation uses adaptive bit-flip attacks as the hardest case of a stored-weight fault.
\end{itemize}
The rest of the paper is organized as follows. Section~\ref{sec:background} explains why the sensitivity distribution matters, through the attacks, the codes and the formats that depend on it. Section~\ref{sec:method} presents the objective that reshapes it. Section~\ref{sec:results} measures the reshaping, checks what each term does and that the damage really falls, and measures the effects on protection, precision and storage; Section~\ref{sec:app} discusses what that means for the systems that would use it and extends the measurement to language models; Section~\ref{sec:related} reviews related work and Section~\ref{sec:concl} concludes.

\section{Background}
\label{sec:background}
This section explains why the sensitivity distribution matters, through the three things that depend on it: the attacks that exploit its tail, the codes that must cover it, and the formats that must store it. It also introduces the baselines.

\paragraph{Bit-flip attacks on stored weights}
Stored weights are reached by Rowhammer-style faults, by low-voltage operation and by hardware or infrastructure Trojans \cite{rahaman2026ispy,rahaman2026evolving}; whatever the vector, the damage depends on which weight is hit. Progressive bit search (PBS) \cite{rakin2019bfa} is the standard adaptive attacker against a quantized network: at each step it ranks stored bits by the gradient of the loss, evaluates the loss after flipping each of the top candidates, commits the worst, and repeats; a handful of flips take ImageNet ResNets to chance. T-BFA \cite{rakin2021tbfa} uses the same search to send one source class to a chosen target class and is reported by its attack success rate. Rowhammer makes such flips practical in DRAM \cite{kim2014rowhammer,yao2020deephammer}, and low-voltage SRAM produces random ones \cite{stutz2021bit}. Both attacks succeed by finding the tail of the sensitivity distribution; in all our experiments the attacker is white-box and sees the weights, the gradients and, where a code is present, the code.

\paragraph{The sensitivity distribution, tolerance and cost}
For a weight stored as an 8-bit two's complement integer with a per-tensor scale, flipping the sign bit changes the integer by 128, which is half of the representable range. Every weight in a layer therefore moves by the same large amount when its sign bit flips, and the damage is decided by the gradient alone. Gradients are heavy-tailed: in a trained ResNet-18 the 1\% of weights with the largest first-order damage account for about 20\% of the total, and the next 8\% for most of the rest; the curves in Figure~\ref{fig:sens} fall by a factor of 10,000 or more between the median weight and the most sensitive one. This non-uniformity is the fact every selective scheme builds on, and it is also what PBS exploits, since the gradient ranking that PBS uses is the same ranking that defines the sensitive set. It is worth being explicit about the two roles the distribution plays. The first is error tolerance: a random flip usually lands in the LOW mass and does little, while an adaptive one is aimed at the tail and does much, so the tolerance of a model to a given number of flips is set by how many weights sit above the damage level at which a flip matters, not by the size of the model. The second is efficiency: a selective code must cover the tail, a mixed-precision format must store it at full width, and a budget of bits or parity buys protection in proportion to how small the tail is. A model whose sensitivity distribution has a short tail is therefore both more tolerant and cheaper to protect, for the same code and the same format. Robust training methods change the distribution as a side effect; the method in Section~\ref{sec:method} changes it on purpose and by a chosen amount.

\paragraph{Selective error correction}
A code over every weight bit stops flips outright but costs 12 to 20\% of model memory in parity, which embedded codes such as E3-CODE \cite{kamineni2026e3} reduce by folding parity into the weight storage. A sensitivity-aware selective code protects selectively: it sorts weights into HIGH, MID and LOW bins by first-order damage, encodes the four most significant bit planes of the HIGH weights with a BCH(127,43) code that corrects $t=12$ errors per 127-bit codeword, and stores the parity in the low-order bit planes of LOW weights, where it costs nothing visible in accuracy. Only four of the eight planes are coded, so the parity is at most $4/8\times 84/43\approx 0.98\%$ of the weight bits for every 1\% of weights covered, and 0.91\% as packed in Table~\ref{tab:ecc}, and it occupies low-order bits of LOW weights rather than extra memory; its cost is a parity budget inside the model, not bits added to it. Defeating one codeword takes more than $t$ errors in that word. In our attack evaluation a flip on a covered bit is modelled as corrected, with no persistent effect; this models an interleaved deployment under the assumption that the attacker cannot accumulate more than $t$ errors in one codeword before it is corrected. The code itself is not simulated, and the guarantee rests on that assumption. In-place zero-space protection \cite{guan2019inplace} and similar schemes make the same trade. For all of them the cost is set by one number, the size of the HIGH bin. The method in the next section is built to make that number small.

\paragraph{Quantization baselines}
For storage we compare against LSQ \cite{esser2020lsq}, which learns the quantization step size by gradient descent and is the standard strong baseline for low-bit uniform quantization, and against KURE \cite{shkolnik2020kure}, which regularizes the kurtosis of the weight distribution so that one model quantizes well at several bit widths. Both are retrained in our pipeline with the same schedule as our own models.

\section{Method}
\label{sec:method}
We first write down what a flip costs, then turn that cost into a training term, show what one gradient step does to it, and finally describe the storage format that a small sensitive set makes affordable.

\paragraph{Sensitivity of a stored weight}
We store weights as INT8 with one scale per tensor: layer $\ell$ has $s_\ell=\max|W_\ell|/127$, and a weight $w$ is kept as the integer $q=\mathrm{round}(w/s_\ell)$. In two's complement, flipping the sign bit of an 8-bit code changes $q$ by 128, so every weight in the layer moves by the same amount $\Delta_\ell=128\,s_\ell$, whatever its own magnitude. The first-order change in the loss produced by that flip is
\begin{equation}
d_i=\Big|\frac{\partial L}{\partial w_i}\Big|\,\Delta_{\ell(i)},
\label{eq:damage}
\end{equation}
which we call the damage, or sensitivity, of weight $i$. We fix two thresholds once, from the baseline model: the HIGH threshold is the damage of the weight at the top 1\% of the baseline distribution and the MID threshold that at the top 9\% (so HIGH holds 1\% of the weights and MID the next 8\%), so the baseline has 1\% HIGH, 8\% MID and 91\% LOW. A trained model is always measured against these fixed thresholds, so ``fewer HIGH weights'' means fewer weights above the baseline's own cut, and the three bins of any model sum to the same total.

\paragraph{Objective}
Let $T$ be the set of the $k=\lceil 0.01N\rceil$ weights with the largest damage. \lean minimizes
\begin{equation}
\begin{aligned}
L_\lambda &= L_{\mathrm{CE}} + \lambda\,P_{\mathrm{LEAN}},\qquad P_{\mathrm{LEAN}}=P_{\mathrm{tail}} + \beta\,P_{\mathrm{clip}},\\
P_{\mathrm{tail}}&=\tfrac1k\textstyle\sum_{i\in T} d_i ,
\end{aligned}
\label{eq:loss}
\end{equation}
where $P_{\mathrm{clip}}=\sum_\ell \mathrm{mean}\big(\mathrm{relu}(|W_\ell|-q_{99}(|W_\ell|))\big)^2$ bounds the INT8 range of each layer, and with it the flip step $\Delta_\ell$; $\beta=100$ in every experiment and is never tuned, so the effective weight of the range term is $100\lambda$ and grows with the knob. The two terms act on the two factors of Eq.~\ref{eq:damage}: $P_{\mathrm{clip}}$ shrinks $\Delta_\ell$ where a layer has outlying weights, which is what sets the sign-flip step, and $P_{\mathrm{tail}}$ lowers $|g_i|$ on the current tail. Section~\ref{sec:results} separates the two; at INT8 the range term does most of the reshaping and the tail term adds to it. $P_{\mathrm{tail}}$ is a function of the gradient, so its gradient is a Hessian-vector product. We compute it by double backpropagation on every fourth step, with the tail term scaled by four on that step so that its average weight over training is $\lambda$; the fine-tune then costs about 1.5 times a plain one. In the double backward pass the step $\Delta_\ell$ (and with it the scale $s_\ell$), the quantile that selects $T$ and the quantile $q_{99}$ are detached, so the second-order term differentiates only through $g$, as in the proposition below. All experiments use ten epochs from the baseline checkpoint with SGD, learning rate 0.002 on CIFAR and 0.0005 on ImageNet, batch size 128 and 32.

\begin{figure}[t]
\centering\includegraphics[width=\columnwidth]{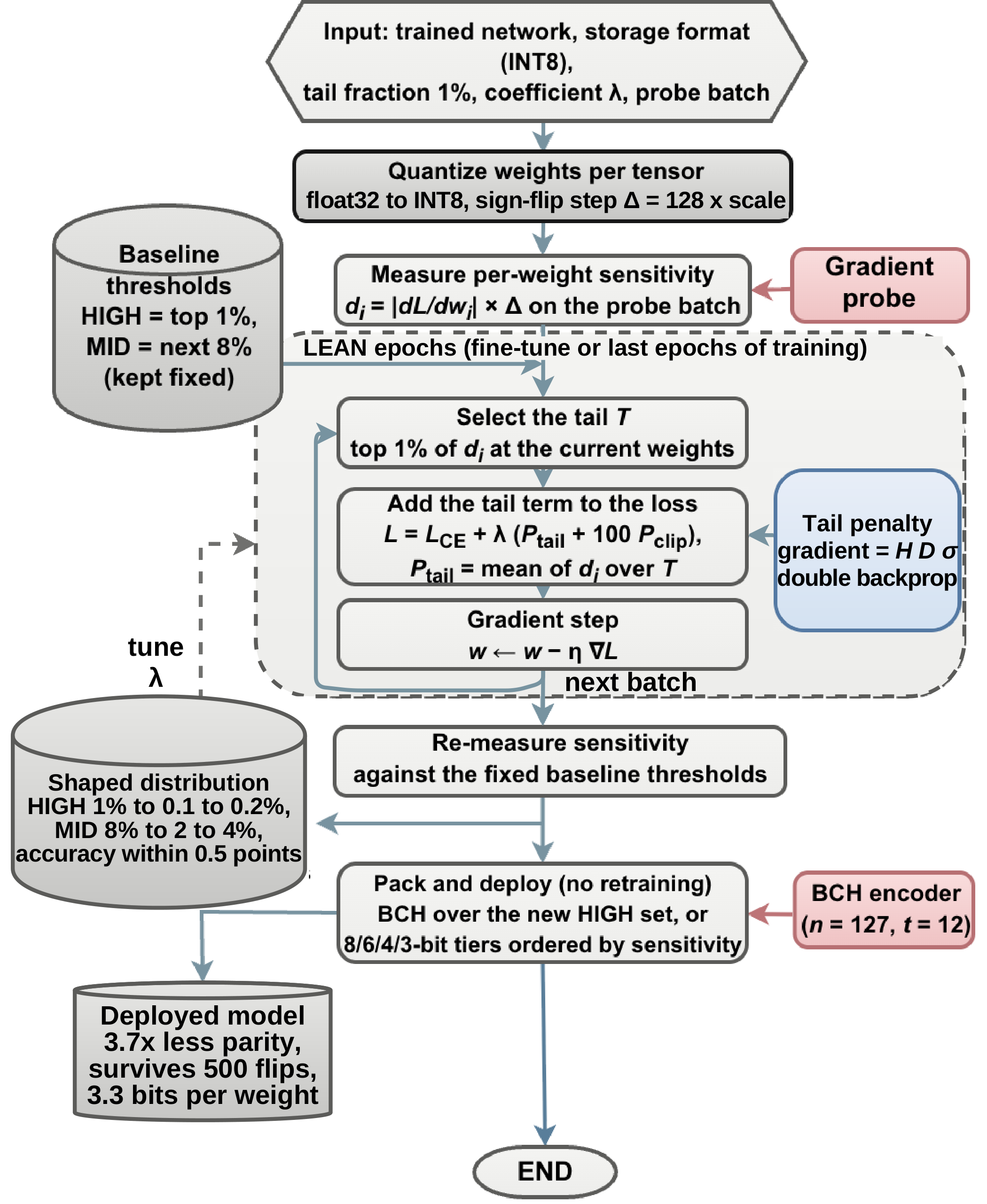}
\caption{The \lean procedure. The trained network is quantized and its sensitivity distribution is measured once, which fixes the HIGH and MID thresholds; ten epochs with the \lean term follow, as a fine-tune or as the last epochs of training, with the tail $T$ re-selected and the tail term applied, through double backpropagation, every fourth step; the model is then re-measured against the fixed thresholds and packed. $\lambda$ is the only knob; the low-bit format uses a separate quantization-aware fine-tune (last paragraph of this section).}
\label{fig:flow}
\end{figure}

Algorithm~\ref{alg:lean} lists the steps, and Figure~\ref{fig:flow} shows the same procedure as a flow, from a trained network to a deployed one. The INT8 pipeline has one \lean stage of ten epochs, a fine-tune or the last epochs of training: the baseline is measured, trained with the \lean term, re-measured against the thresholds fixed from the baseline, and packed, either with a code over a fixed budget (the top 1\% by damage, as in all attack tables) or over everything above $\tau_H$ (the threshold mode of the parity experiment). The low-bit format of the last paragraph of this section is a second deployment and uses a second, quantization-aware fine-tune.

\begin{algorithm}[t]
\caption{\lean: shaping the sensitivity distribution}
\label{alg:lean}
\begin{algorithmic}[1]
\Require trained $w$; sign-flip step $\Delta_\ell$ of the storage format; $\lambda$; probe batch $B$
\State $d_i \gets |\partial L_{\mathrm{CE}}/\partial w_i|\,\Delta_{\ell(i)}$ on $B$ \Comment{baseline}
\State $\tau_H, \tau_M \gets$ top-1\% and top-9\% quantiles of $d$ \Comment{fixed}
\For{each fine-tuning step with batch $(x,y)$}
  \State $g \gets \nabla_w L_{\mathrm{CE}}(x,y)$
  \State $u \gets 0$
  \If{step mod 4 $= 0$} \Comment{graph retained}
    \State $d_i \gets |g_i|\,\Delta_{\ell(i)}$;\; $T \gets$ top $k=\lceil 0.01N\rceil$ of $d$
    \State $P_{\mathrm{tail}} \gets \frac1k\sum_{i\in T} d_i$ \Comment{$\Delta_\ell$, $T$ detached}
    \State $u \gets 4\,\nabla_w P_{\mathrm{tail}}$ by double backprop \Comment{$=4HD\sigma$}
  \EndIf
  \State $w \gets w - \eta\,\Big[\,g + \lambda\,\big(u + \beta\,\nabla_w P_{\mathrm{clip}}\big)\Big]$ \Comment{Eq.~\ref{eq:loss}}
\EndFor
\State re-measure $d$ on $B$; shares above $\tau_H$, $\tau_M$ \Comment{HIGH, MID}
\State pack: BCH over the top 1\% by $d$ (fixed budget) or over $\{i: d_i>\tau_H\}$ (threshold)
\end{algorithmic}
\end{algorithm}

\begin{proposition}[Observation]
Let $g=\nabla_w L_{\mathrm{CE}}$, $H=\nabla^2_w L_{\mathrm{CE}}$, $D=\mathrm{diag}(\Delta_{\ell(i)})$ and $\sigma_i=\mathrm{sign}(g_i)\,\mathbf{1}[i\in T]/k$. Hold $T$, the signs of $g$ on $T$ and $D$ fixed. A gradient step of size $\eta$ on $\lambda P_{\mathrm{tail}}$ alone changes the tail sensitivity by
$\Delta P_{\mathrm{tail}} = -\eta\lambda\,\|HD\sigma\|^2 + O(\eta^2)$.
\end{proposition}
With $T$ and the signs held fixed over the step, $P_{\mathrm{tail}}=\sigma^\top D g$, so $\nabla_w P_{\mathrm{tail}}=HD\sigma$; the step is $w\leftarrow w-\eta\lambda HD\sigma$, and the first-order change of $P_{\mathrm{tail}}$ is $-\eta\lambda\,(HD\sigma)^\top(HD\sigma)$.
This is the descent lemma applied to $P_{\mathrm{tail}}$ and we claim nothing more for it; it is local to the tail-penalty step, and $T$ can change between steps. Its use is to say what the tail term can and cannot do: it moves $w$ along the Hessian rows of the tail weights and lowers their gradients, which is the quantity a gradient-guided attacker reads, and a detached proxy such as $|w_i|$ or a frozen $|g_i|$ has no $H$ term and cannot do that. It does not act on $\Delta_\ell$, which is the other factor of the damage and, as Section~\ref{sec:results} shows, the larger lever at INT8. Whether lowering the gradient also lowers the real damage of a flip is an empirical question, and we answer it with real flips rather than with the proxy.

\paragraph{Sensitivity-ordered low-bit storage}
The proposition covers the training side. On the deployment side, a cheaper code and a longer survival need no change to the storage format, but the third consequence, a smaller model, does. Once HIGH and MID are small, a mixed format can afford to store them well. We keep the HIGH tier at 8 bits with a per-tensor scale, the MID tier at 6 bits per channel, fill a 4-bit tier with the next most sensitive weights until an average-bit budget is met, and store the remaining weights at 3 bits per channel. This is a second, ten-epoch quantization-aware fine-tune, started from the \lean INT8 checkpoint, that carries the same $\lambda$ term, now evaluated with each weight's own deployed flip step $2^{b-1}s$; the tiers are refreshed every 50 of its steps. The bits-per-weight figures we report count the weight payload only; the per-channel scales and the tier map are not included, and code parity is added only where stated. The map is the larger of the two: tiers are assigned per weight, so an uncompressed map costs 2 bits per weight, an entropy-coded one about 1 bit at the tier shares of Table~\ref{tab:storage}, and a block-wise assignment would cost a fraction of a bit; we leave its layout to the deployment and compare payloads. Two details decide whether this low-bit model can be attacked cheaply. The scale of the 6/4/3-bit tiers is set by the 0.99 quantile of the tier rather than its maximum, so that a sign flip in the 3-bit tier moves a weight 2 to 4 times less than it otherwise would; and tiers are ranked by deployed-format damage, so that the 8-bit tier is the most damaging 1\% of the model that is actually stored, not of the floating-point model it came from.

\section{Results}
\label{sec:results}
The experiments are in two parts. The first part establishes that the procedure reshapes the distribution on every model, by an amount that $\lambda$ controls, without hurting accuracy. The second part measures the consequences of the reshaped distribution for bit-flip protection, for precision and storage, and for the low-bit model's own resilience. Language models are in Section~\ref{sec:app}.

\begin{figure}[t]
\centering\includegraphics[width=\columnwidth]{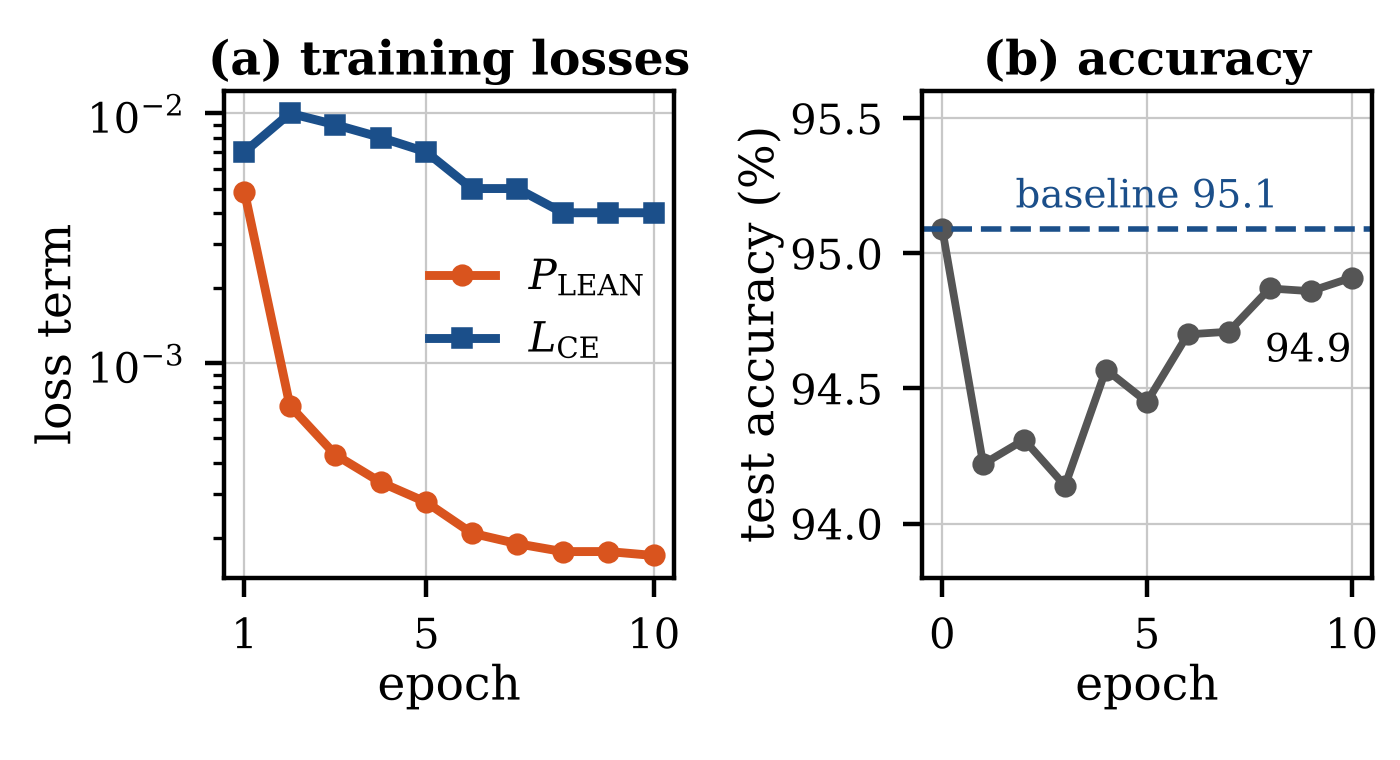}
\caption{One \lean fine-tune, ResNet-18 CIFAR-10, $\lambda=40$: (a) the regularizer falls 28 times while the cross-entropy stays flat; (b) test accuracy dips after the first epoch and recovers to within 0.2 points of the baseline.}
\label{fig:loss}
\end{figure}

\begin{figure*}[!t]
\centering\includegraphics[width=0.97\textwidth]{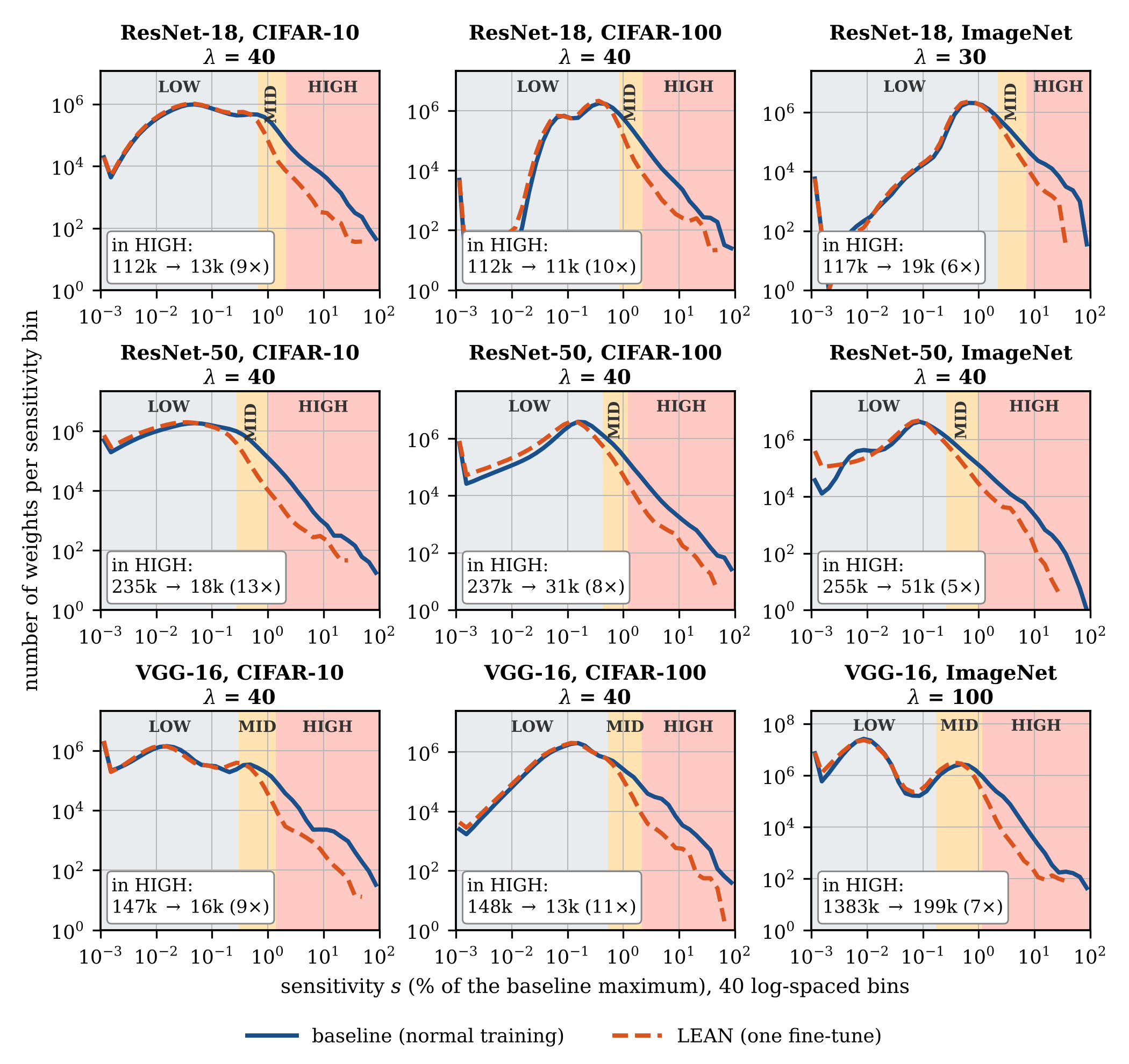}
\caption{Raw weight-sensitivity distribution of the nine models: number of weights per sensitivity bin (40 log-spaced bins of $d_i$ as a percentage of the baseline maximum; the first bin collects every weight below $10^{-3}$), baseline solid, \lean dashed, with the baseline LOW, MID and HIGH bands. Figure~\ref{fig:sens} shows the same data as a cumulative distribution. The bulk of the distribution is unchanged; the HIGH tail moves left into the MID and LOW bands.}
\label{fig:sens_raw}
\end{figure*}

\begin{figure*}[!t]
\centering\includegraphics[width=0.97\textwidth]{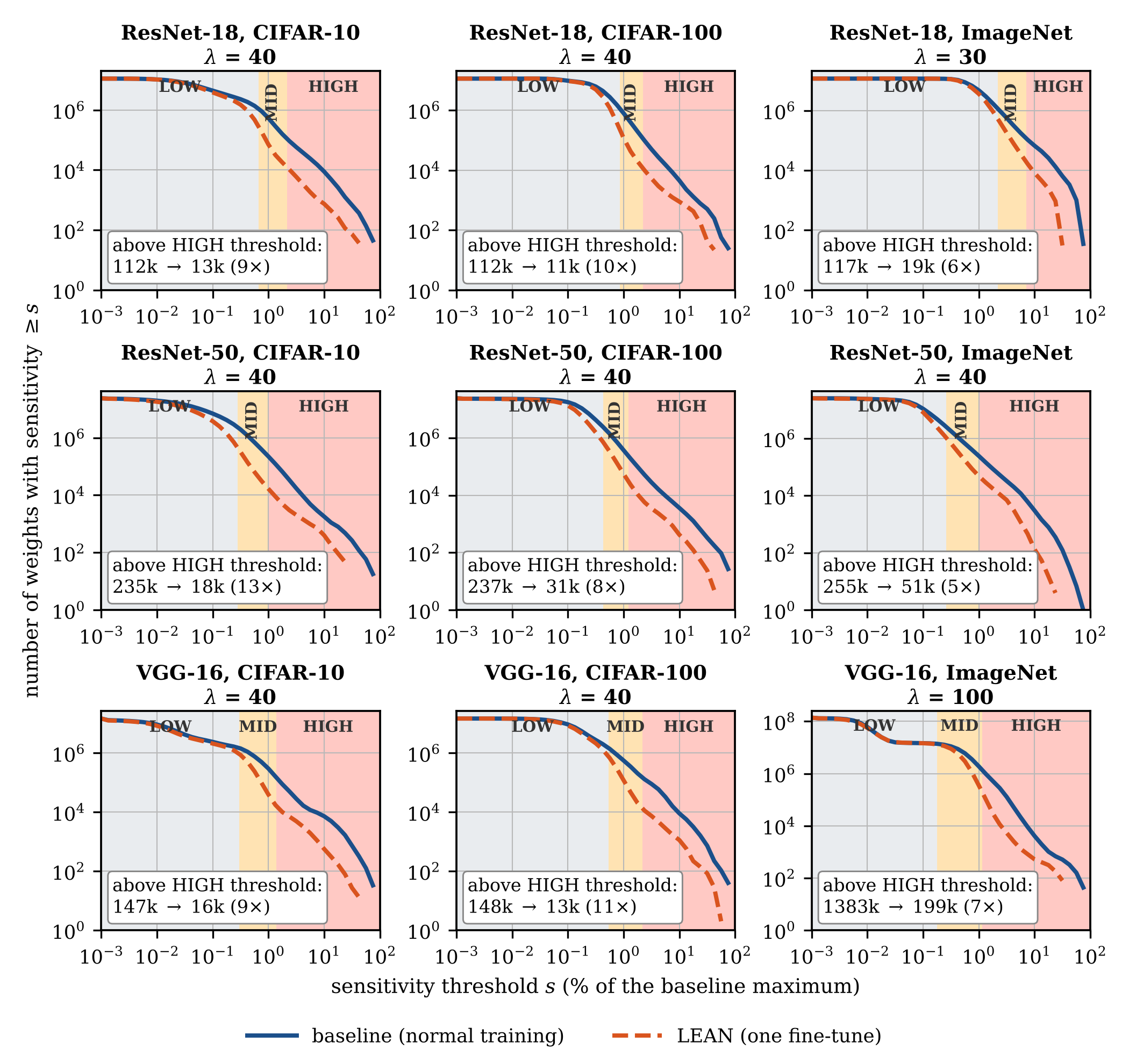}
\caption{Cumulative weight-sensitivity distribution of the nine models, counted from the tail (number of weights whose sensitivity is at least $s$), on log-log axes; baseline solid, \lean dashed. The shaded bands are the baseline LOW, MID and HIGH bins. The \lean curve separates from the baseline at the MID boundary and stays about one decade lower across the HIGH band. The box in each panel gives the count above the HIGH threshold before and after.}
\label{fig:sens}
\end{figure*}

\paragraph{Experimental setup}
We use ResNet-18, ResNet-50 \cite{he2016resnet} and VGG-16 with batch normalization \cite{simonyan2015vgg} on CIFAR-10, CIFAR-100 \cite{krizhevsky2009cifar} and ImageNet \cite{deng2009imagenet}; ResNet-34 is added in the storage and low-bit experiments (Tables~\ref{tab:cost}, \ref{tab:lowbit} and \ref{tab:lowbit_imnet}) as a fourth deployment target. CIFAR baselines are trained from scratch. ImageNet baselines are the torchvision pretrained models, and every ImageNet fine-tune and evaluation in this paper uses one fixed random split of the 50k validation set (seed 0), 40k images for fine-tuning and 10k held out for evaluation; ImageNet accuracies are therefore on this 10k held-out set and are not directly comparable to published 50k-set numbers. Every \lean INT8 model is a single ten-epoch fine-tune from its baseline with the objective of Eq.~\ref{eq:loss} (Figure~\ref{fig:loss} shows one such run); the published defenses and the quantization baselines are fine-tuned with the same schedule. Unless a table says otherwise, each entry is one training run and one attack run with a fixed seed; Table~\ref{tab:storage} uses three seeds, and the ResNet-18 CIFAR-10 $\lambda=40$ fine-tune was repeated with three seeds: clean accuracy 94.84 $\pm$ 0.06, HIGH 0.113 $\pm$ 0.003\%, MID 1.84 $\pm$ 0.17\%, and with the 1\% code every seed holds 94\% through 300 PBS flips while two of three hold 94\% at 500 (the third falls to chance there). The tail fraction is 1\% throughout; 0.5\% and 2\% give HIGH 0.115\% and 0.118\% at 94.7\% and 94.9\% accuracy, so the result is not sensitive to it. The protection experiments instantiate the selective code with the layout of Section~\ref{sec:background}, but \lean does not depend on it: any scheme whose cost grows with the number of high-sensitivity weights benefits from the same tail reduction. Sensitivity is measured by Eq.~\ref{eq:damage} on 20 batches of 128 images, against thresholds fixed from the baseline: the thresholds used during training come from training batches, and the reported distributions, counts and shares (Table~\ref{tab:sens}, Figures~\ref{fig:sens_raw} and \ref{fig:sens}) from held-out batches. Attacks are real PBS with 50 candidate bits per step and two attack batches of 128 images, reported as top-1 accuracy on the held-out set after $n$ committed flips; T-BFA is reported as accuracy and attack success rate after 30 flips. Clean accuracy is top-1 on the held-out set. All experiments were run in PyTorch on NVIDIA RTX A6000 GPUs (48 GB of memory each), one GPU per run; a ten-epoch \lean fine-tune takes about 9 minutes for ResNet-18 on CIFAR-10 and 1 to 3 hours on ImageNet, and a 500-flip PBS attack takes 25 minutes on CIFAR and 1 to 3 hours on ImageNet.

\paragraph{The distribution moves on every model}
Table~\ref{tab:sens} counts the weights above the baseline HIGH threshold before and after \lean on nine models; Figure~\ref{fig:sens_raw} shows the raw distribution and Figure~\ref{fig:sens} the cumulative distribution, counted from the tail. The count falls 5 to 13 times. The share of the total first-order damage carried by the top 1\% of weights falls 1.4 to 2.1 times on all nine models (Table~\ref{tab:sens}, from 9 to 28\% down to 5 to 18\%); a uniform rescaling of the flip steps would leave this share unchanged, so the distribution is reshaped, not only scaled. The same share computed from the gradients alone, with no flip step, is unchanged by \lean (8.2 against 8.7\% on ResNet-18 CIFAR-10, and alike elsewhere), so the reshaping acts through the per-layer flip steps, which the range term lowers where the tail lives, while the tail term's contribution shows in the real-flip damage of Table~\ref{tab:ablate} rather than in this share. Both figures use logarithmic axes, which is why the change looks modest at first sight: the bulk of the distribution spans six decades of count and is untouched, while the tail that moves is one decade lower, a factor of 10 in the number of weights that the code has to cover. The clean-accuracy change is at most 0.5 points (ResNet-50 on CIFAR-100) and is usually 0.1 or less; on ImageNet ResNet-18 it is zero. The curves make clear that the change is not confined to the very top of the distribution: the \lean curve leaves the baseline curve at the MID boundary and runs about one decade below it through the whole HIGH band, so the entire tail has moved to lower sensitivity. Where the weights go is also visible: since every weight is counted somewhere, a lower curve at the MID line is the same statement as a larger LOW bin, and on ResNet-18 CIFAR-10 LOW grows from 91\% to about 92\% while HIGH drops from 1\% to 0.11\%.

\begin{table}[t]
\caption{HIGH-bin weights before and after \lean, the share of total first-order damage carried by the top 1\% of weights (\%, scale-free, measured on 20 held-out batches; the ablation of Table~\ref{tab:ablate} measures the same quantity on its own probe batches, 20.4 $\to$ 11.7\%) and clean accuracy (\%), each as base $\to$ \lean; $\lambda=40$ unless noted.}
\label{tab:sens}
\centering\scriptsize\setlength{\tabcolsep}{2pt}
\begin{tabular}{llrrrrr}
\toprule
data & model & base & \lean & $\times$ & top-1\% share & acc. \\
\midrule
C10 & ResNet-18 & 112k & 13k (0.11\%) & 9 & 19.5 $\to$ 10.8 & 95.1 $\to$ 94.9 \\
C10 & ResNet-50 & 235k & 18k (0.08\%) & 13 & 16.0 $\to$ 10.9 & 95.0 $\to$ 94.8 \\
C10 & VGG-16 & 147k & 16k (0.11\%) & 9 & 28.3 $\to$ 17.9 & 92.9 $\to$ 92.8 \\
C100 & ResNet-18 & 112k & 11k (0.10\%) & 10 & 9.1 $\to$ 5.1 & 77.2 $\to$ 77.1 \\
C100 & ResNet-50 & 237k & 31k (0.13\%) & 8 & 10.2 $\to$ 7.1 & 75.9 $\to$ 75.4 \\
C100 & VGG-16 & 148k & 13k (0.09\%) & 11 & 18.3 $\to$ 8.6 & 72.3 $\to$ 72.3 \\
ImNet & ResNet-18 ($\lambda$30) & 117k & 19k (0.16\%) & 6 & 11.6 $\to$ 6.2 & 69.9 $\to$ 69.9 \\
ImNet & ResNet-50 & 255k & 51k (0.20\%) & 5 & 14.9 $\to$ 9.2 & 80.7 $\to$ 80.5 \\
ImNet & VGG-16 ($\lambda$100) & 1383k & 199k (0.14\%) & 7 & 26.0 $\to$ 18.3 & 73.4 $\to$ 73.3 \\
\bottomrule
\end{tabular}
\end{table}

\begin{figure*}[!t]
\centering\includegraphics[width=0.97\textwidth]{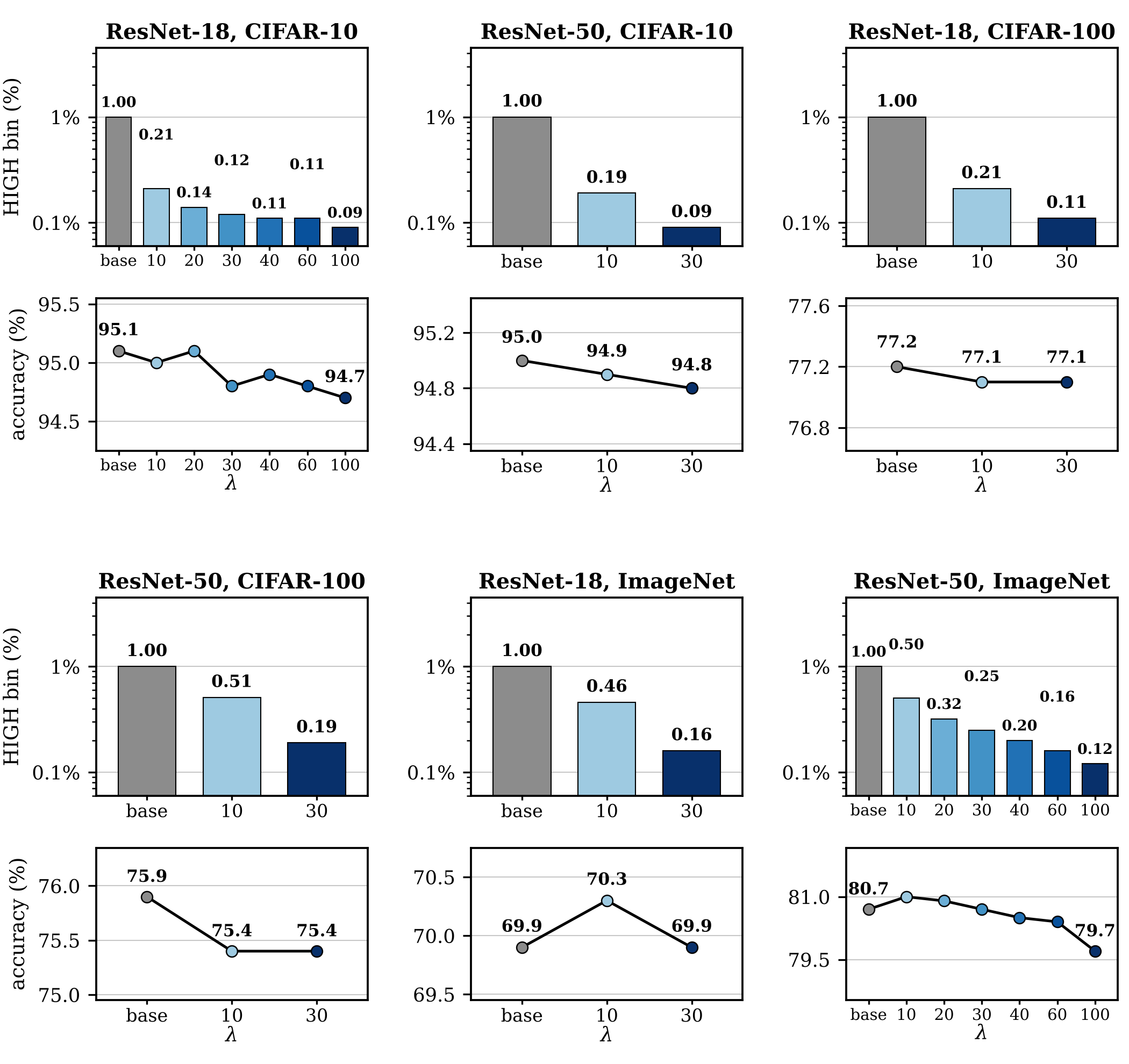}
\caption{The knob. HIGH-bin share (bars, log scale) and clean accuracy (line) against $\lambda$ for six models, at the values trained for the sweep. The HIGH bin falls monotonically and saturates near 0.1\%; accuracy stays within 0.5 points up to $\lambda=40$ and drops at $\lambda=100$.}
\label{fig:knob}
\end{figure*}

\paragraph{$\lambda$ is a knob}
Table~\ref{tab:sens} uses one $\lambda$ per model; Figure~\ref{fig:knob} plots the HIGH share and the clean accuracy against $\lambda$ for every value we trained on six models. The HIGH share falls monotonically with $\lambda$ and saturates near 0.1\% of the weights; accuracy stays within 0.5 points of the baseline up to $\lambda=40$ and begins to fall at $\lambda=100$ (ResNet-50 ImageNet goes from 80.7 to 79.7). We therefore use $\lambda=40$ on CIFAR and $\lambda=30$ to $40$ on ImageNet unless stated otherwise. The two ImageNet models are the ones that shrink least (5 to 6 times); their sensitive sets are the largest to begin with and need the strongest penalty.

\begin{figure*}[!t]
\centering\includegraphics[width=0.97\textwidth]{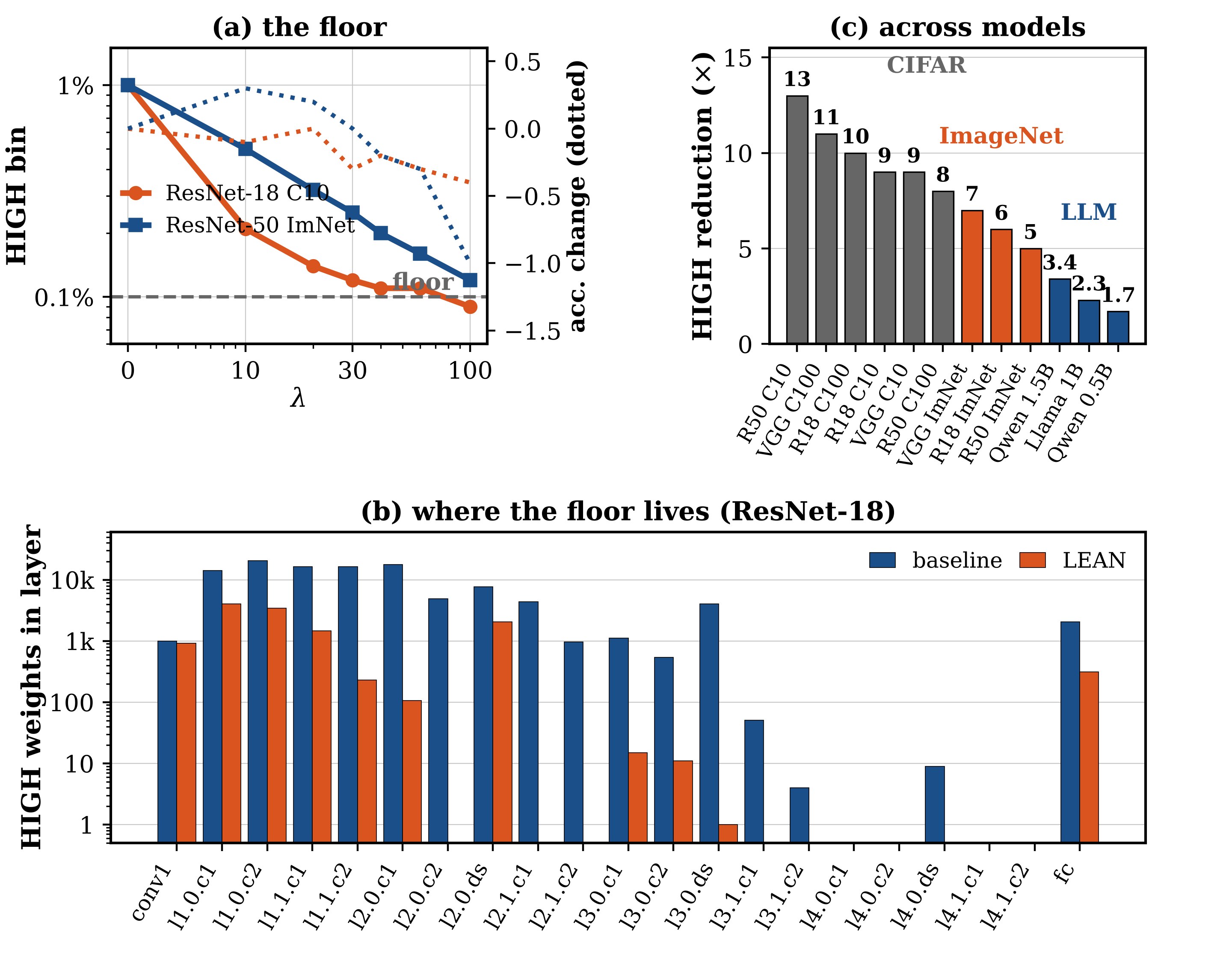}
\caption{The limit of the reshaping. (a) HIGH share against $\lambda$ (solid, left axis) and change in clean accuracy (dotted, right axis) for ResNet-18 CIFAR-10 and ResNet-50 ImageNet; the share flattens near 0.1\% while accuracy begins to fall. (b) HIGH weights per layer before and after \lean, ResNet-18 CIFAR-10, $\lambda=40$: the deep layers empty, the first convolution and first stage keep most of what remains. (c) HIGH reduction factor for the twelve models, largest to smallest: CIFAR, then ImageNet, then the language models.}
\label{fig:floor}
\end{figure*}

\paragraph{What sets the floor}
The knob does not go to zero, and Figure~\ref{fig:floor} shows why. The damage of a weight is $|g_i|\Delta_\ell$, so the tail can be lowered only as far as its two factors allow. The range term shrinks $\Delta_\ell$ down to the level of a layer's ordinary weights and no further without clipping them, so the flip step of a layer falls by at most its outlier ratio,
\begin{equation}
\Delta_\ell \ge \tfrac{128}{127}\,q_{99}(|W_\ell|),\qquad \frac{\Delta_\ell^{\mathrm{before}}}{\Delta_\ell^{\mathrm{after}}}\le \rho_\ell=\frac{\max|W_\ell|}{q_{99}(|W_\ell|)},
\label{eq:floor_delta}
\end{equation}
which is a large gain in a layer with outlying weights and none in a layer without them. The tail term lowers the top-$k$ gradients at a speed set by the curvature term $\|HD\sigma\|$ of the Observation, and because the set $T$ is re-selected every step, what it does in practice is flatten the top of the distribution into a plateau rather than remove it (Figure~\ref{fig:sens}). Where the two stop is the joint optimum $\lambda\nabla_w P_{\mathrm{LEAN}}=-\nabla_w L_{\mathrm{CE}}$, beyond which every further reduction of the tail is paid for in task loss. On ResNet-18 CIFAR-10 the HIGH bin falls monotonically with $\lambda$ and flattens near 0.1\% from $\lambda=30$ on, while accuracy begins to drop (Figure~\ref{fig:floor}a); the penalty itself plateaus by epoch 6 to 8 of the fine-tune (Figure~\ref{fig:loss}). Where the remaining weights sit is instructive (Figure~\ref{fig:floor}b): the deep layers empty completely, and four fifths of the 12.6k weights still above the threshold are in the first convolution and the first stage, small layers that every input passes through, whose weights have no outliers to clip (the first convolution has $\max|W|/q_{99}|W|=1.7$, the lowest in the network, and keeps 92\% of its HIGH weights) and whose gradients cannot be lowered without changing what the network computes. The floor is therefore a property of the model, set by its per-layer outlier ratios $\rho_\ell$ (Eq.~\ref{eq:floor_delta}) and by the curvature of the loss around its early weights, not by $\lambda$ or the number of epochs; $\lambda$ chooses how far toward it to go. Across models the reduction is largest where there are few weights and large outliers, 9 to 13 times on CIFAR, and smallest where many weights sit near the threshold, 5 to 7 times on ImageNet and 1.7 to 3.4 times on the language models (Figure~\ref{fig:floor}c).

\begin{figure*}[!t]
\centering\includegraphics[width=0.78\textwidth]{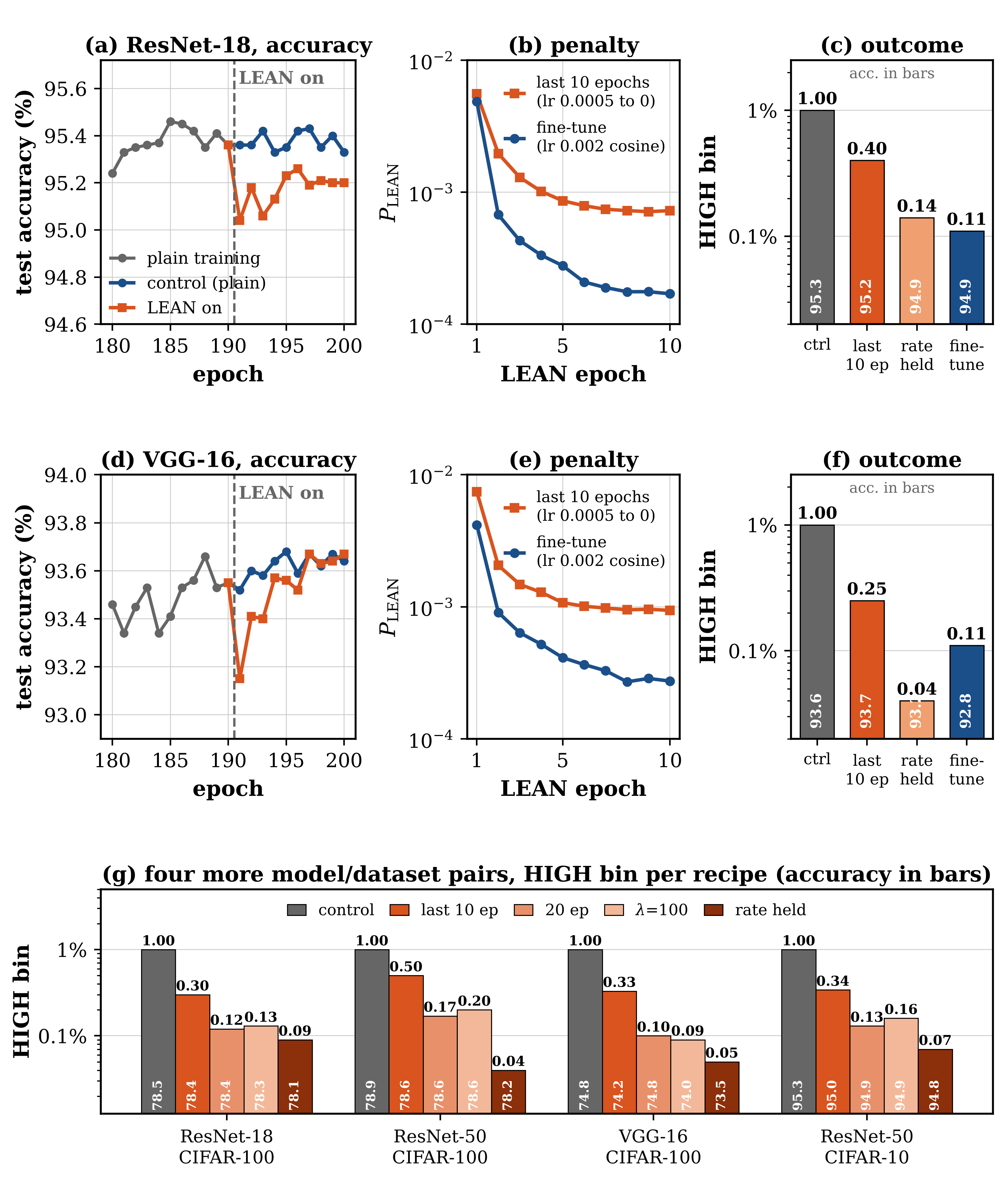}
\caption{\lean inside training, $\lambda=40$; ResNet-18 CIFAR-10 in the top row, VGG-16 CIFAR-10 below. (a, d) Test accuracy for epochs 180 to 200 of a 200-epoch run; at epoch 190 the run is forked into a plain control and a branch with the \lean term on. (b, e) The penalty over the ten \lean epochs inside training (learning rate 0.0005 to 0) and in the fine-tune (0.002, cosine). (c, f) HIGH share and accuracy of the control, the last-epochs model, the last-epochs model with the rate held at 0.002, and the fine-tune. (g) HIGH share per recipe on four more pairs, accuracy inside the bars.}
\label{fig:inset}
\end{figure*}

\paragraph{Fine-tune or last epochs of training}
\lean is a loss term, so it can be switched on inside normal training instead of run as a separate fine-tune. Figure~\ref{fig:inset} and Tables~\ref{tab:inset} and \ref{tab:inset2} test this on six model/dataset pairs: one 200-epoch run from scratch per pair, forked at epoch 190 into a plain control and branches with the term on for the last epochs. Switched on as is, in the last ten epochs of the cosine schedule, the term works but reaches less far than the fine-tune, HIGH 0.25 to 0.50\% against 0.11\%, and the penalty is still falling at epoch 200 (Figure~\ref{fig:inset}b). The cause is the learning rate, which in those epochs is 0.0005 falling to zero, a quarter of the fine-tune's. Holding it at the fine-tune's 0.002 for the same ten epochs reaches the floor on every pair, HIGH 0.04 to 0.14\% and MID 0.7 to 2.9\%, at the fine-tune's accuracy; twenty epochs or $\lambda=100$ at the dying rate get halfway, HIGH 0.05 to 0.20\%. Reaching the floor costs 0.1 to 0.7 points except on VGG-16 CIFAR-100 (1.3 points, where a smaller $\lambda$ would be used), stopping halfway costs 0.1 or nothing. Under attack the routes line up in the same order (Table~\ref{tab:def}, bottom block, ResNet-18 CIFAR-10 with the 1\% code): the control breaks at 200 flips, the last-epochs model at 300, the rate-held model holds 94\% at 300 and breaks at 500; the targeted attack with the code stays at 3 to 6\% success for all of them, as for the fine-tune. Giving the last ten epochs the fine-tune's full schedule, 0.002 decaying to zero (rate held, cos.), closes the remaining gap: 93\% at 500 flips at 95.0\% clean accuracy, the fine-tune's protection at a slightly higher accuracy, and on VGG-16 the same schedule reaches HIGH 0.04\% at no accuracy cost. The decay matters more than the last few hundredths of HIGH: twenty epochs at a constant 0.002 (rate held, 20 ep.) leave a smaller tail by count (0.13\%, MID 0.6\%) yet break at 400 flips, because the undecayed rate leaves the surviving weights less settled. Both routes are the same technique; we use the fine-tune because it gives the term that schedule by construction and because the ImageNet and language-model baselines are pretrained checkpoints with no training of ours to switch it on in.

\begin{table}[t]
\caption{\lean in the last epochs of a 200-epoch run, CIFAR-10; HIGH and MID against the thresholds of the matching control of the same run. The fine-tune rows are the paper's models, measured against their own baseline.}
\label{tab:inset}
\centering\scriptsize\setlength{\tabcolsep}{2.5pt}
\begin{tabular}{lrlr|rrr|rrr}
\toprule
& & & & \multicolumn{3}{c|}{ResNet-18} & \multicolumn{3}{c}{VGG-16} \\
recipe & ep. & rate & $\lambda$ & acc & HIGH & MID & acc & HIGH & MID \\
\midrule
control & 10 & 0.0005$\to$0 & 0 & 95.3 & 1.00 & 8.0 & 93.6 & 1.00 & 8.0 \\
last epochs & 10 & 0.0005$\to$0 & 40 & 95.2 & 0.40 & 6.4 & 93.7 & 0.25 & 6.9 \\
longer & 20 & 0.0006$\to$0 & 40 & 95.3 & 0.20 & 3.9 & 93.5 & 0.06 & 6.2 \\
stronger & 10 & 0.0005$\to$0 & 100 & 95.0 & 0.19 & 3.9 & 93.2 & 0.05 & 5.3 \\
rate held & 10 & 0.002 & 40 & 94.9 & \textbf{0.14} & \textbf{0.9} & 93.2 & \textbf{0.04} & \textbf{2.9} \\
rate held, cos. & 10 & 0.002$\to$0 & 40 & 95.0 & 0.16 & 2.2 & 93.6 & \textbf{0.04} & 4.2 \\
rate held, 20 ep. & 20 & 0.002 & 40 & 95.1 & 0.13 & 0.6 & 92.7 & 0.04 & 2.0 \\
fine-tune & 10 & 0.002 cos. & 40 & 94.9 & \textbf{0.11} & 2.0 & 92.8 & 0.11 & 4.8 \\
\bottomrule
\end{tabular}
\end{table}

\begin{table}[t]
\caption{The same recipes on four more pairs (C-10, C-100: CIFAR), $\lambda=40$ unless stated, ten epochs unless stated. HIGH and MID in \% of weights.}
\label{tab:inset2}
\centering\scriptsize\setlength{\tabcolsep}{1.6pt}
\begin{tabular}{llrrr|llrrr}
\toprule
model & recipe & acc & HIGH & MID & model & recipe & acc & HIGH & MID \\
\midrule
ResNet-18 & control & 78.5 & 1.00 & 8.0 & ResNet-50 & control & 78.9 & 1.00 & 8.0 \\
C-100 & last 10 & 78.4 & 0.30 & 4.9 & C-100 & last 10 & 78.6 & 0.50 & 5.6 \\
 & 20 ep. & 78.4 & 0.12 & 2.4 & & 20 ep. & 78.6 & 0.17 & 3.1 \\
 & $\lambda$=100 & 78.3 & 0.13 & 2.6 & & $\lambda$=100 & 78.6 & 0.20 & 3.3 \\
 & rate held & 78.1 & \textbf{0.09} & \textbf{0.7} & & rate held & 78.2 & \textbf{0.04} & \textbf{1.2} \\
\midrule
VGG-16 & control & 74.8 & 1.00 & 8.0 & ResNet-50 & control & 95.3 & 1.00 & 8.0 \\
C-100 & last 10 & 74.2 & 0.33 & 6.6 & C-10 & last 10 & 95.0 & 0.34 & 4.7 \\
 & 20 ep. & 74.8 & 0.10 & 4.2 & & 20 ep. & 94.9 & 0.13 & 2.8 \\
 & $\lambda$=100 & 74.0 & 0.09 & 3.8 & & $\lambda$=100 & 94.9 & 0.16 & 3.1 \\
 & rate held & 73.5 & \textbf{0.05} & \textbf{0.9} & & rate held & 94.8 & \textbf{0.07} & \textbf{0.9} \\
\bottomrule
\end{tabular}
\end{table}

\paragraph{Which term does the work, and is the damage real}
Two questions have to be settled before the consequences are measured. The first is whether the proxy $d_i=|g_i|\Delta_{\ell(i)}$ can be gamed: an attacker ranks weights by the gradient, and a term that lowers the gradient could hide weights without making them safer. The second is which of the two terms of Eq.~\ref{eq:loss} moves the distribution. Table~\ref{tab:ablate} answers both on ResNet-18 CIFAR-10. HIGH and the top-1\% damage share are from the proxy. For the first question we flip real sign bits one at a time, in the INT8 model, for the 1000 weights the proxy ranks highest and for 1000 random weights, and record the true loss increase; the table gives the mean over the 100 worst weights found (worst 100) and over the baseline's own top 1000 re-flipped inside each fine-tuned model (base top), and random weights cause no measurable change in any model. $\Delta$ ratio is the mean per-layer flip step relative to the baseline. In the \lean model the worst 100 weights cost 0.009 per flip against 0.031 in the baseline, and the baseline's top 1000 cost 0.0014 against 0.0069, so the damage really falls; the proxy's share of total damage held by the top 1\%, which does not depend on the scale of $\Delta_\ell$, falls from 20\% to 12\%. For the second, the range term alone reaches almost the same point (HIGH 0.13\%, worst-100 damage 0.012, $\Delta_\ell$ halved on average) at no accuracy cost, and the tail term on top of it lowers the damage a further 25\% (0.012 to 0.009, and 0.0019 to 0.0014 on the fixed set). The tail term alone does what the first question feared: at $\lambda=400$ it reports a 3 times smaller HIGH bin while the real damage of the worst weights is unchanged and accuracy drops a point. At INT8, then, the lever is the flip step, which the range term controls, and the tail term is a useful addition that must not be used on its own. The last row is the plain gradient-norm penalty, the squared norm of the weight gradient, under the same schedule, the nearest generic regularizer: at $\lambda=0.01$, the largest value that keeps accuracy within a point, it leaves HIGH at 0.64\% and the real damage of the worst weights at 0.026 against 0.031, and at $\lambda=0.03$ accuracy falls to 88\% with the tail unchanged; without the storage-format weighting and the tail selection the gradient is smoothed everywhere and the tail is not where the pressure goes.

\begin{table}[t]
\caption{Ablation and real-flip check, ResNet-18 CIFAR-10, $\lambda=40$ unless noted. Columns are defined in the text.}
\label{tab:ablate}
\centering\scriptsize\setlength{\tabcolsep}{2.5pt}
\begin{tabular}{lrrrrrr}
\toprule
model & acc & HIGH & top-1\% & worst 100 & base top & $\Delta$ ratio \\
\midrule
baseline & 95.1 & 1.00\% & 20.4\% & 0.031 & 0.0069 & 1.00 \\
$P_{\mathrm{tail}}$ only & 94.5 & 0.87\% & 19.3\% & 0.030 & 0.0061 & 0.99 \\
$P_{\mathrm{tail}}$ only, $\lambda$=400\! & 94.1 & 0.35\% & 16.3\% & 0.031 & 0.0062 & 1.07 \\
$P_{\mathrm{clip}}$ only & 95.1 & 0.13\% & 12.8\% & 0.012 & 0.0019 & 0.51 \\
gradient norm, $\lambda$=0.01 & 94.5 & 0.64\% & 19.5\% & 0.026 & 0.0058 & 1.02 \\
\lean (both) & 94.9 & 0.11\% & 11.7\% & \textbf{0.009} & \textbf{0.0014} & 0.51 \\
\bottomrule
\end{tabular}
\end{table}

\paragraph{Effect on bit-flip protection}
The first consequence of a smaller tail is that a small code becomes enough. We attack the INT8 model with real progressive bit search \cite{rakin2019bfa} over all layers and protect it, where stated, with the selective BCH(127,43) code ($t=12$) on the top four bit planes of the HIGH weights. Table~\ref{tab:def} places \lean next to four published robust-training defenses retrained in the same pipeline: piecewise clustering \cite{he2020pc}, WCAT \cite{chitsaz2023wcat}, Forget-and-Rewire \cite{nazari2024far} and Hessian-aware training \cite{pranto2025hessian}. Without a code every method, \lean included, is at chance within 50 to 100 flips: an attacker that may flip anything always finds the next most damaging weight. With the same 1\% code the methods separate, and the HIGH column gives the first reason: the defenses leave 0.89 to 1.05\% of the weights above the threshold (WCAT 0.17\%), so a 1\% code is exactly full and PBS finds the next weight. A small tail is necessary but not sufficient; what also counts is how damaging the weights just below the code are, the worst-100 column of Table~\ref{tab:ablate}: the range-only model, WCAT and the rate-held 20-epoch model all leave 0.13 to 0.17\% HIGH yet end at 49, 75 and 16\% at 500 flips. In what follows a model breaks at the first flip count at which its accuracy falls below 90\%. The normal model breaks at 200 flips, piecewise clustering and Hessian-aware training at 200, Forget-and-Rewire at 300, WCAT at 500 (75\%); \lean at $\lambda=40$ holds 94\% at 500 (in two of three training seeds; the third holds 94\% at 300 and breaks at 500, the experimental setup of Section~\ref{sec:results}), and the $\lambda$ sweep ($\lambda=10$ failing at 500, $\lambda=30$ holding 90\%) shows the dependence directly. The ablation rows settle what each term buys under attack: the range term alone (HIGH 0.13\%) holds at 200 and breaks at 300, short of WCAT, and the tail term alone breaks at 200 like the normal model, or at 500 when pushed to $\lambda=400$; only the two together pass WCAT. The plain gradient-norm fine-tune breaks at 200. The code removes the weights the attacker wants, and $\lambda$ makes sure that nothing of comparable sensitivity is left behind them. On ImageNet ResNet-18 with the 1\% code the \lean model holds 83\% at 100 flips and 72\% at 200 against 69\% and 41\% for the normal model. Table~\ref{tab:tbfa} repeats the exercise with the targeted attack \cite{rakin2021tbfa} on all nine models, 30 flips, three class pairs each. CIFAR uses the held-out test set. ImageNet uses all 50 validation images of each class (25 for the attack, 25 for the success rate), 40 of which were in the \lean fine-tuning split while the normal model is the torchvision checkpoint that never saw them, so the overlap favours the \lean side and the ImageNet rows should be read with that in mind; the CIFAR rows have no such overlap. Accuracy on the non-source classes stays within one point of clean on CIFAR; on ImageNet VGG-16 without a code it falls to 22 to 27\%. The same rule holds, and it also shows where the rule stops. Where the targeted attack goes through weights that the mixed-data sensitivity ranks highly (the CIFAR-10 models and ImageNet VGG-16, whose classifier holds most of its tail), \lean with the code ends it: 1 to 5\% success against 30 to 93\% for the normal model with the same code. Where the attack lives in the last layer's row for one of 100 or 1000 classes, weights whose mixed-data sensitivity is modest, \lean lowers the success rate on CIFAR-100 and changes nothing on the ImageNet ResNets. The reason is that \lean reshapes the sensitivity it is trained on, the damage under the mixed-class training objective, and robustness does not transfer by itself to a class-conditional sensitivity that this objective does not rank. A tier chosen by the maximum damage over target classes closes that route for the normal model too; it is a packing-time choice that does not separate \lean from the baseline, so we report the plain ranking and count the ImageNet ResNets as a limit of the mixed-data objective rather than of the code.

\begin{table}[t]
\caption{ResNet-18 CIFAR-10, INT8, accuracy (\%) after $n$ progressive bit-search flips, without and with the same 1\% BCH code; HIGH in \%. Clean accuracy 95. Bottom block: the models trained with \lean in the last epochs.}
\label{tab:def}
\centering\scriptsize\setlength{\tabcolsep}{2.3pt}
\begin{tabular}{lr|rrrr|rrrr}
\toprule
& & \multicolumn{4}{c|}{no code} & \multicolumn{4}{c}{with 1\% BCH code} \\
method & HIGH & 10 & 25 & 50 & 100 & 100 & 200 & 300 & 500 \\
\midrule
normal & 1.00 & 41 & 11 & 10 & 10 & 93 & 26 & 17 & 10 \\
PC \cite{he2020pc} & 0.89 & 44 & 11 & 10 & 10 & 95 & 33 & 22 & 14 \\
Hessian \cite{pranto2025hessian} & 1.01 & 44 & 11 & 10 & 10 & 95 & 31 & 22 & 14 \\
FaR \cite{nazari2024far} & 1.05 & 16 & 10 & 10 & 10 & 95 & 95 & 45 & 12 \\
WCAT \cite{chitsaz2023wcat} & 0.17 & 80 & 48 & 18 & 11 & 95 & 95 & 94 & 75 \\
\lean $\lambda=10$ & 0.21 & 77 & 30 & 10 & 10 & 95 & 94 & 93 & 16 \\
\lean $\lambda=30$ & 0.12 & 67 & 25 & 12 & 11 & 95 & 94 & 94 & 90 \\
\lean $\lambda=40$ & 0.11 & 81 & 36 & 10 & 10 & \textbf{95} & \textbf{95} & \textbf{95} & \textbf{94} \\
\midrule
\multicolumn{10}{l}{ablations and gradient-norm baseline (Table~\ref{tab:ablate}):} \\
$P_{\mathrm{clip}}$ only & 0.13 & 79 & 25 & 10 & 10 & 95 & 95 & 77 & 49 \\
$P_{\mathrm{tail}}$ only & 0.87 & 45 & 11 & 10 & 10 & 94 & 43 & 34 & 10 \\
$P_{\mathrm{tail}}$, $\lambda$=400 & 0.35 & 26 & 10 & 10 & 10 & 94 & 93 & 91 & 10 \\
gradient norm & 0.64 & 47 & 10 & 10 & 10 & 94 & 14 & 15 & 11 \\
\midrule
\multicolumn{10}{l}{\lean inside training (Table~\ref{tab:inset}):} \\
control & 1.00 & 86 & 11 & 10 & 10 & 95 & 93 & 78 & 10 \\
last epochs & 0.40 & 94 & 90 & 15 & 10 & 95 & 95 & 31 & 10 \\
rate held & 0.14 & 94 & 77 & 26 & 10 & 95 & 95 & 94 & 11 \\
rate held, cos. & 0.16 & 94 & 65 & 18 & 10 & 95 & 95 & 95 & \textbf{93} \\
rate held, 20 ep. & 0.13 & 95 & 91 & 53 & 13 & 95 & 94 & 93 & 16 \\
\bottomrule
\end{tabular}
\end{table}

\begin{table}[t]
\caption{T-BFA, 30-flip budget, mean attack success rate (\%) over three class pairs, without and with the 1\% code; \lean at the $\lambda$ of Table~\ref{tab:sens}. Data and caveats in the text.}
\label{tab:tbfa}
\centering\footnotesize\setlength{\tabcolsep}{4pt}
\begin{tabular}{lrr|rr}
\toprule
& \multicolumn{2}{c|}{no code} & \multicolumn{2}{c}{with 1\% BCH code} \\
model & normal & \lean & normal & \lean \\
\midrule
C10 ResNet-18 & 95.4 & 44.6 & 92.7 & \textbf{3.9} \\
C10 ResNet-50 & 95.8 & 25.3 & 86.1 & \textbf{5.3} \\
C10 VGG-16 & 56.2 & 7.3 & 30.2 & \textbf{5.4} \\
C100 ResNet-18 & 96.0 & 75.3 & 96.7 & 70.7 \\
C100 ResNet-50 & 96.7 & 82.7 & 85.3 & 55.3 \\
C100 VGG-16 & 92.7 & 32.0 & 93.3 & 28.7 \\
ImageNet ResNet-18 & 96.0 & 96.0 & 90.7 & 92.0 \\
ImageNet ResNet-50 & 98.7 & 96.0 & 97.3 & 96.0 \\
ImageNet VGG-16 & 41.3 & 21.3 & 73.3 & \textbf{1.3} \\
\bottomrule
\end{tabular}
\end{table}

\begin{table}[t]
\caption{Parity for a fixed guarantee, ResNet-18 CIFAR-10, INT8 with the selective code; coverage in \% of weights, parity in \% of weight bits, accuracy (\%) at 6\% BER on the covered and on the uncovered weights, and accuracy drop (points) at 25 flips. The 6\% BER figure on the uncovered weights is not reported for the coverage sweep, since at one seed it depends on which layers the random errors land in.}
\label{tab:ecc}
\centering\scriptsize\setlength{\tabcolsep}{3pt}
\begin{tabular}{lrrrrr}
\toprule
model & cov. & parity & BER prot. & BER unprot. & PBS-25 drop \\
\midrule
normal & 20.0\% & 18.0 & 95.0 & 88.0 & 0.00 \\
normal & 19.2\% & 17.6 & 94.9 & -- & 0.48 \\
normal & 16.8\% & 15.4 & 95.0 & -- & 0.48 \\
normal & 15.2\% & 13.9 & 95.0 & -- & 0.52 \\
normal & 12.5\% & 11.5 & 95.0 & 88.3 & 0.52 \\
\lean $\lambda$=10 & 11.9\% & 10.9 & 95.0 & 90.6 & 0.00 \\
\lean $\lambda$=10 & 5.3\% & \textbf{4.8} & 95.0 & 94.1 & 0.00 \\
\lean $\lambda$=40 & 5.3\% & 4.8 & 94.9 & 93.9 & 0.16 \\
\bottomrule
\end{tabular}
\end{table}

\begin{figure*}[!t]
\centering\includegraphics[width=0.78\textwidth]{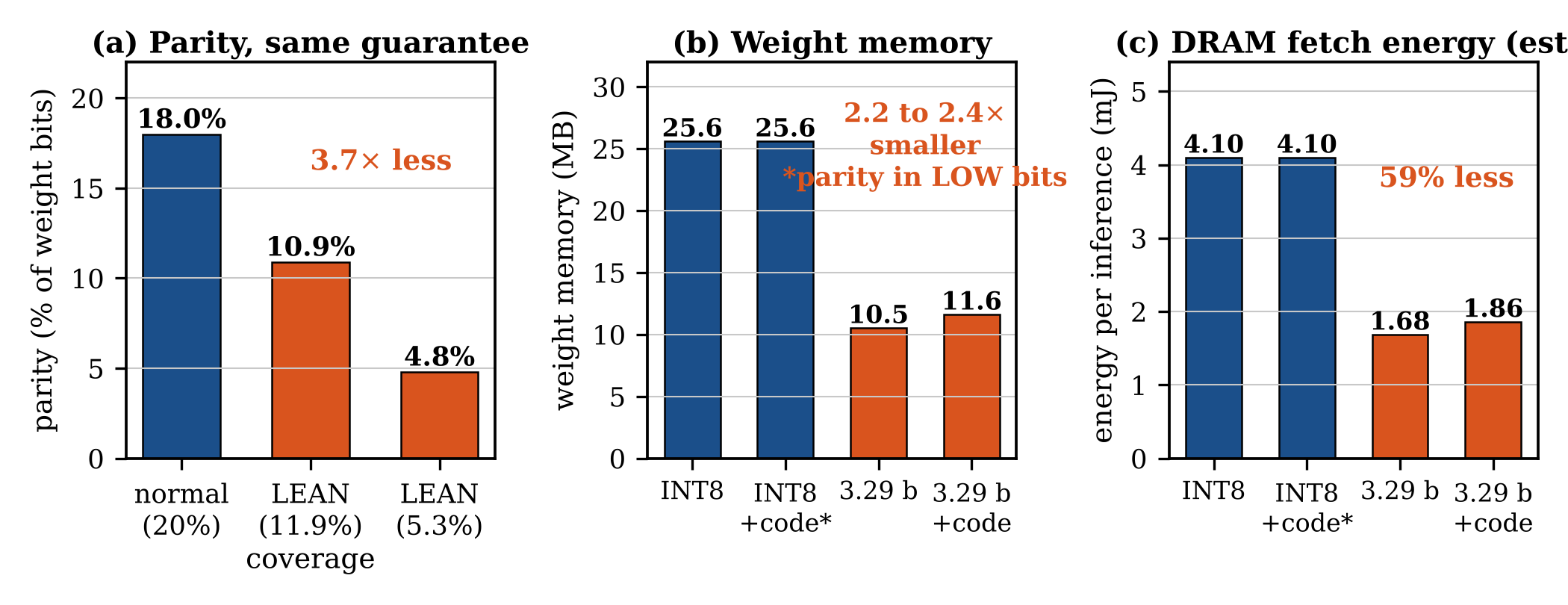}
\caption{Cost of protection, storage and weight fetch before and after \lean. (a) Parity for the same protection guarantee (Table~\ref{tab:ecc}), ResNet-18 CIFAR-10, measured; coverage in brackets. (b) Weight memory of ImageNet ResNet-50 at INT8, where the 1\% code's parity is embedded in LOW weights, and in the tiered 3.29-bit format without and with the 1\% BCH and $t=19$ sign-plane codes (3.63 bits per weight); the 4.09-bit ImageNet hardening mode of Table~\ref{tab:lowbit_imnet} is not shown. (c) DRAM weight-fetch energy per inference with the 45 nm access energies of Han et al.\ \cite{han2016eie}; on-chip SRAM scales identically (32 to 13 $\mu$J).}
\label{fig:storage}
\end{figure*}

\paragraph{Effect on parity, precision, storage and energy}
The second consequence of a smaller tail is cost: less parity, fewer bits and less energy for the same model. Figure~\ref{fig:storage} collects the three cost numbers. Protection first. The guarantee is fixed in advance and has a random and an adaptive part: 95\% accuracy at a 6\% random bit error rate on the protected weights, and no accuracy drop at 25 progressive flips (Table~\ref{tab:ecc}; the code is BCH $t=12$ over four planes with the parity embedded in LOW weights, parity in \% of weight bits; BER prot.\ and BER unprot.\ are the accuracy when the 6\% error rate hits the covered weights with the code on, and the uncovered weights with no code). The 6\% rate is a stress level, far above the $10^{-3}$ to $10^{-2}$ of low-voltage SRAM, chosen so that the guarantee is conservative; the 25-flip point sizes the parity, and the 500-flip survival of Table~\ref{tab:def} is the separate, stronger result for the $\lambda=40$ model with the 1\% code. The top 1\% by damage is not enough for the normal ResNet-18, because PBS moves on to the weights just below the cut, which are nearly as damaging; it needs the code on 20\% of the weights, which is 18\% of the weight bits as parity: coverages of 12.5, 15.2, 16.8 and 19.2\% all lose about 0.5 points at 25 flips, so the minimum lies between 19.2 and 20\%. The \lean model with $\lambda=10$ meets the same guarantee with the code on 5.3\% of the weights, 4.8\% parity, because the weights below its cut are far less damaging; the parity is 3.7 times smaller. We use $\lambda=10$ in this table because it is the smallest $\lambda$ that meets the zero-drop guarantee; the paper's $\lambda=40$ model at the same 5.3\% coverage misses it by 0.16 points at 25 flips, and loses one point when the 6\% error rate hits its uncovered weights. The same table gives the random-fault side: with no code on the uncovered weights, a 6\% bit error rate there costs the normal model seven points at 20\% coverage and the \lean model four points at 11.9\% and one point at 5.3\%. Since the HIGH bin shrinks with negligible change in accuracy (Table~\ref{tab:sens}), the parity budget of any selective scheme shrinks with it.

Storage next. Table~\ref{tab:storage} compares formats for ImageNet ResNet-50 at exactly matched (4.00) and comparable (3.00 against 3.29) bits per weight under identical ten-epoch training. At a uniform 3 bits our quantization-aware training reaches 74.3\%, the same as LSQ \cite{esser2020lsq} (74.4\%), so $\lambda$ by itself is not a better quantizer. The gain decomposes as follows: sensitivity-ordered tiering at 3.29 bits per weight adds 2.4 points over the uniform 3-bit model, and $\lambda$ adds 0.4 more, consistently over three seeds. Most of the improvement is therefore from the ordering; what \lean contributes is a smaller 8-bit and 6-bit tier, which is what makes the ordering affordable at this budget, plus the last 0.4 points. At an exactly matched 4.00 bits the tiered \lean model is 1.2 points above LSQ and 3.7 above KURE \cite{shkolnik2020kure}. In bytes, the 25.6 MB INT8 model becomes 10.5 MB of weight payload, or 11.6 MB with the codes of the next paragraph (at 3 bits per weight there are no LOW low-order bits to hide parity in, so there it is counted), not counting scales and the tier map.

Energy last. This is an analytical estimate, not a measurement: fewer bits per weight means fewer bits cross the memory interface, and we scale the weight traffic by published access energies. Taking the 45 nm access energies of Han et al.\ \cite{han2016eie}, 640 pJ for a 32-bit DRAM access and 5 pJ for a 32-bit SRAM access, the weight fetch for one ResNet-50 inference falls from 4.1 mJ to 1.7 mJ from DRAM and from 32 to 13 $\mu$J from on-chip SRAM, an estimated 59\% less weight-fetch energy in either case; tier lookup, scale and metadata fetch and decoder activity are outside this estimate. The same reduction applies to the SRAM weight buffer an accelerator must provision, which shrinks 2.2 to 2.4 times. Table~\ref{tab:cost} gives the same two numbers for every model in the paper: INT8 with the 1\% code stays at 8 bits per weight because the parity is embedded in LOW weights; the low-bit format with both codes is 3.63 bits per weight (CIFAR, Table~\ref{tab:lowbit}) or 4.09 with the 3\% tier and the $t=79$ code (ImageNet, Table~\ref{tab:lowbit_imnet}, given for the CIFAR models for comparison); the CIFAR and ImageNet ResNets of the same depth differ only in the classifier, so one row covers both; decoder and tier-lookup energy are not included. The ratio is fixed by the format (2.2 times at 3.63 bits per weight, 2.0 times at 4.09, with the INT8 parity embedded), so what changes from model to model is the absolute saving, from 1 mJ per inference on a CIFAR ResNet-18 to 12.5 mJ on ImageNet VGG-16.

\begin{table}[t]
\caption{ImageNet ResNet-50 top-1 (\%) at matched (4.00) and comparable (3.00 vs 3.29) bits per weight; tiered rows are mean of three seeds.}
\label{tab:storage}
\centering\footnotesize
\begin{tabular}{lrr}
\toprule
format & bits/w & top-1 \\
\midrule
INT8 & 8.00 & 80.7 \\
LSQ uniform \cite{esser2020lsq} & 4.00 & 78.4 \\
KURE uniform \cite{shkolnik2020kure} & 4.00 & 75.9 \\
tiered, $\lambda=0$ & 4.00 & 78.9 \\
tiered, \lean $\lambda=40$ & 4.00 & \textbf{79.6} \\
\midrule
LSQ uniform & 3.00 & 74.4 \\
KURE uniform & 3.00 & 73.2 \\
ours, uniform QAT & 3.00 & 74.3 \\
tiered, $\lambda=0$ & 3.29 & 76.8 \\
tiered, \lean $\lambda=40$ & 3.29 & \textbf{77.2} \\
\bottomrule
\end{tabular}
\end{table}

\paragraph{Effect on the low-bit model's own resilience}
The INT8 model and the 3.29-bit model are different deployments; the natural question is whether the low-bit one can be protected by the 1\% code as well. It cannot, and the reason is itself a statement about the distribution: at 3 bits per channel almost every weight is one sign flip away from a change of its full magnitude, so the tail the code must cover is the whole model (Table~\ref{tab:lowbit}, left block, chance by 100 flips). Three changes, none of them retraining, close the gap, at a price in clean accuracy of about five points on ImageNet (Table~\ref{tab:lowbit_imnet}, 71.8\% against 77.2\% for the storage-only format of Table~\ref{tab:storage}), which is the cost of the smaller scales and the deployed-format ranking. The 6/4/3-bit scales are set by the 0.99 quantile instead of the maximum, which shrinks every low-bit flip 2 to 4 times; the tiers are ranked by deployed-format damage; and a long interleaved BCH code over the sign-bit plane (1023 sign bits per codeword, $t=19$, 0.19 bits per weight) spreads each codeword over random addresses, so an adaptive attacker must spend $t+1$ flips to gain one. With these, the CIFAR models hold 100 to 500 flips within 2 to 7 points of clean at 3.63 bits per weight (Table~\ref{tab:lowbit}, right block). ImageNet needs more: $t=79$ and a 3\% tier obtained by promoting the 2\% most damaging uncovered weights at packing time, at 4.09 bits per weight, after which VGG-16 (72.5\% clean after promotion) is unchanged by 500 flips and ResNet-34 loses one point, while ResNet-18 and ResNet-50 lose 14 and 13 (Table~\ref{tab:lowbit_imnet}). The general lesson for the code is that correction over neighbouring words does not help against an adaptive flipper, since neighbouring weights are as useful to it as one; interleaving is what makes the attacker's extra flips worthless.

\begin{table}[t]
\caption{CIFAR models at 3.29 bits per weight under progressive bit search, accuracy (\%): BCH on the 1\% tier only (3.45 b/w) and with the interleaved sign-plane code (3.63 b/w).}
\label{tab:lowbit}
\centering\footnotesize\setlength{\tabcolsep}{3.5pt}
\begin{tabular}{lrrrrrrr}
\toprule
& \multicolumn{3}{c}{BCH only} & \multicolumn{4}{c}{+ sign-plane code} \\
\cmidrule(lr){2-4}\cmidrule(lr){5-8}
model & 25 & 50 & 100 & 100 & 200 & 300 & 500 \\
\midrule
C10 ResNet-18 & 91 & 68 & 14 & 94 & 94 & 94 & 90 \\
C10 ResNet-34 & 93 & 89 & 73 & 94 & 94 & 94 & 93 \\
C10 ResNet-50 & 82 & 63 & 28 & 94 & 93 & 92 & 90 \\
C10 VGG-16 & 83 & 65 & 27 & 93 & 92 & 91 & 86 \\
C100 ResNet-18 & 67 & 56 & 11 & 75 & 72 & 71 & 67 \\
C100 ResNet-34 & 72 & 51 & 16 & 76 & 76 & 75 & 70 \\
C100 ResNet-50 & 40 & 12 & 4 & 73 & 66 & 54 & 38 \\
C100 VGG-16 & 50 & 31 & 11 & 70 & 64 & 59 & 49 \\
\bottomrule
\end{tabular}
\end{table}

\begin{table}[t]
\caption{ImageNet models at 3.29 bits per weight under progressive bit search, accuracy (\%) after 100 / 200 / 500 flips, with the sign-plane code at $t=19$ (3.63 b/w) or $t=79$ (3.68), and with the 3\% tier and $t=79$ (4.09). No retraining.}
\label{tab:lowbit_imnet}
\centering\footnotesize\setlength{\tabcolsep}{3pt}
\begin{tabular}{lrrrrrrrrrr}
\toprule
& & \multicolumn{3}{c}{1\% tier, $t$=19} & \multicolumn{3}{c}{1\% tier, $t$=79} & \multicolumn{3}{c}{3\% tier, $t$=79} \\
\cmidrule(lr){3-5}\cmidrule(lr){6-8}\cmidrule(lr){9-11}
model & clean & 100 & 200 & 500 & 100 & 200 & 500 & 100 & 200 & 500 \\
\midrule
ResNet-18 & 67.7 & 54 & 27 & 1 & 63 & 56 & 25 & 67 & 64 & 54 \\
ResNet-34 & 71.1 & 66 & 58 & 22 & 70 & 69 & 64 & 71 & 71 & \textbf{70} \\
ResNet-50 & 71.8 & 8 & 1 & 0 & 48 & 24 & 2 & 69 & 68 & 59 \\
VGG-16 & 72.1 & 65 & 59 & 39 & 71 & 69 & 62 & 73 & 73 & \textbf{73} \\
\bottomrule
\end{tabular}
\end{table}

\section{Applications}
\label{sec:app}
This section says what a reshaped distribution means for the systems that would use it, and extends the measurement to language models.

\begin{table}[t]
\caption{Weight memory (MB) and estimated DRAM weight-fetch energy per inference (mJ) for the deployed formats; formats and assumptions in the text. $^\dagger$Millions of weights.}
\label{tab:cost}
\centering\footnotesize\setlength{\tabcolsep}{3.5pt}
\begin{tabular}{lrrrrrrr}
\toprule
& & \multicolumn{2}{c}{INT8 (+code)} & \multicolumn{2}{c}{3.63 b/w} & \multicolumn{2}{c}{4.09 b/w} \\
\cmidrule(lr){3-4}\cmidrule(lr){5-6}\cmidrule(lr){7-8}
model & Mw$^\dagger$ & MB & mJ & MB & mJ & MB & mJ \\
\midrule
CIFAR ResNet-18 & 11.2 & 11.2 & 1.79 & 5.1 & 0.81 & 5.7 & 0.92 \\
CIFAR ResNet-34 & 21.3 & 21.3 & 3.41 & 9.7 & 1.55 & 10.9 & 1.74 \\
CIFAR ResNet-50 & 23.6 & 23.6 & 3.78 & 10.7 & 1.71 & 12.1 & 1.93 \\
CIFAR VGG-16 & 14.7 & 14.7 & 2.35 & 6.7 & 1.07 & 7.5 & 1.20 \\
ImageNet ResNet-18 & 11.7 & 11.7 & 1.87 & 5.3 & 0.85 & 6.0 & 0.96 \\
ImageNet ResNet-34 & 21.8 & 21.8 & 3.49 & 9.9 & 1.58 & 11.1 & 1.78 \\
ImageNet ResNet-50 & 25.6 & 25.6 & 4.10 & 11.6 & 1.86 & 13.1 & 2.09 \\
ImageNet VGG-16 & 138.4 & 138.4 & 22.14 & 62.8 & 10.05 & 70.8 & 11.32 \\
\bottomrule
\end{tabular}
\end{table}

\paragraph{Accelerators and deployment}
A selective code hides its parity in the low-order bits of LOW weights, so its cost is the number of HIGH weights times the code rate and nothing else. \lean reduces that number 5 to 13 times at training time without touching the memory system: the decoder, the bin table and the parity embedding are reused unchanged, and the only difference the hardware sees is a shorter list of protected addresses. Two deployments follow from the two fine-tunes. The \lean INT8 checkpoint with the 1\% code is a security mode, surviving 300 flips in every seed and 500 in two of three, with its parity embedded in LOW weights at no added memory; the quantization-aware checkpoint in the 8/6/4/3-bit format is a storage mode, 2.4 times smaller in weight payload and, by the access-energy model of Section~\ref{sec:results}, an estimated 59\% cheaper to fetch, and with the sign-plane code it is a security mode too at 3.63 to 4.09 bits per weight. The two checkpoints share the baseline, the thresholds and the $\lambda$ term; a device picks the mode that fits its memory.

\begin{table}[t]
\caption{\lean on language models: HIGH and MID shares against the control's thresholds, and WikiText-2 perplexity at fp32 and after per-channel INT4 (lower is better).}
\label{tab:llm}
\centering\footnotesize\setlength{\tabcolsep}{4pt}
\begin{tabular}{llrrrr}
\toprule
model & $\lambda$ & HIGH & MID & ppl (fp32) & ppl (INT4) \\
\midrule
Qwen2-0.5B & 0 (control) & 1.00\% & 8.0\% & 13.86 & 21.1 \\
Qwen2-0.5B & 10 & 0.58\% & 4.9\% & 14.51 & 19.3 \\
Llama-3.2-1B & 0 (control) & 1.00\% & 8.0\% & 11.47 & 21.4 \\
Llama-3.2-1B & 3 & \textbf{0.43\%} & 4.1\% & 12.09 & \textbf{18.9} \\
Llama-3.2-1B & 10 & \textbf{0.19\%} & 2.1\% & 13.96 & 19.3 \\
Qwen2-1.5B & 0 (control) & 1.00\% & 8.0\% & 10.78 & 13.3 \\
Qwen2-1.5B & 3 & \textbf{0.29\%} & 2.6\% & 12.90 & 14.4 \\
\bottomrule
\end{tabular}
\end{table}

\begin{table}[t]
\caption{Preliminary: progressive sign-bit attack on INT8 language models, perplexity after $n$ flips, without and with a 1\% code by each model's own ranking; ``dead'' is above $10^{6}$.}
\label{tab:llmflip}
\centering\footnotesize\setlength{\tabcolsep}{4pt}
\begin{tabular}{llrrrrr}
\toprule
model & code & clean & 10 & 20 & 50 & 100 \\
\midrule
Llama-3.2-1B control & none & 12.2 & dead & dead & dead & dead \\
Llama-3.2-1B \lean $\lambda$=3 & none & 12.6 & dead & dead & dead & dead \\
Llama-3.2-1B control & 1\% & 12.2 & \textbf{12.5} & \textbf{14.4} & 122 & 210 \\
Llama-3.2-1B \lean $\lambda$=3 & 1\% & 12.6 & 20.8 & 28.5 & \textbf{41.9} & \textbf{61.4} \\
\midrule
Qwen2-0.5B control & none & 14.3 & dead & dead & dead & dead \\
Qwen2-0.5B \lean $\lambda$=10 & none & 14.7 & dead & dead & dead & dead \\
Qwen2-0.5B control & 1\% & 14.3 & 29.5 & dead & dead & dead \\
Qwen2-0.5B \lean $\lambda$=10 & 1\% & 14.7 & \textbf{15.0} & \textbf{115} & dead & dead \\
\midrule
Qwen2-1.5B control & none & 11.0 & dead & dead & dead & dead \\
Qwen2-1.5B \lean $\lambda$=3 & none & 13.0 & dead & dead & dead & dead \\
Qwen2-1.5B control & 1\% & 11.0 & \textbf{11.0} & \textbf{11.2} & dead & dead \\
Qwen2-1.5B \lean $\lambda$=3 & 1\% & 13.0 & 13.1 & 16.3 & dead & dead \\
\bottomrule
\end{tabular}
\end{table}

\begin{figure*}[!t]
\centering\includegraphics[width=0.74\textwidth]{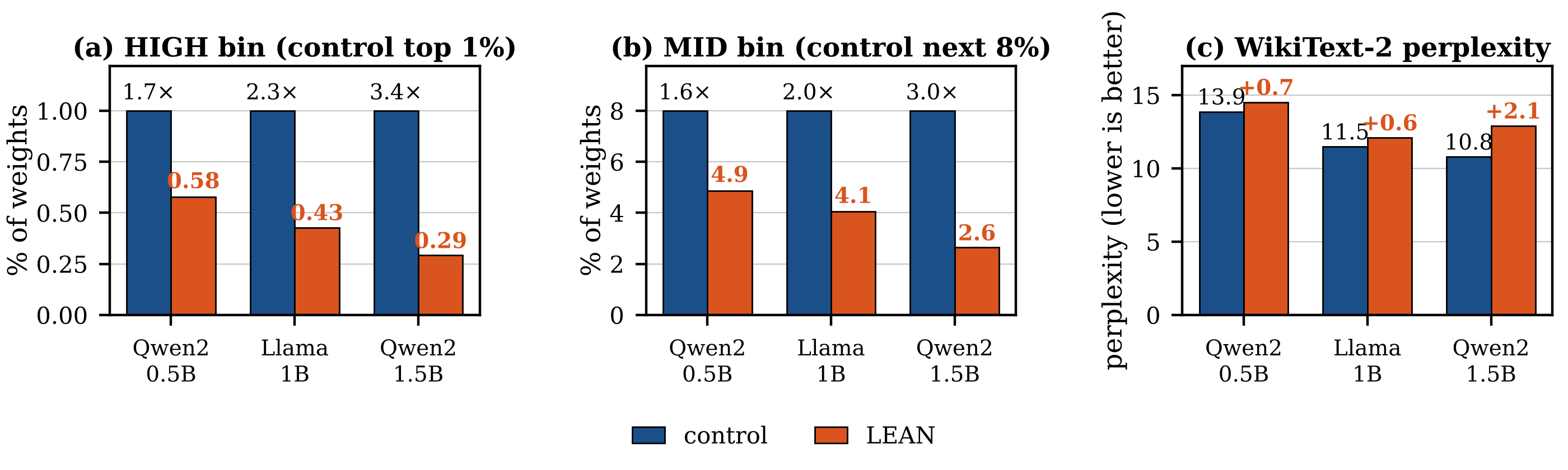}
\caption{\lean on three language models (Table~\ref{tab:llm} in bars): (a) HIGH and (b) MID shares against the control's thresholds, with the reduction factor; (c) WikiText-2 perplexity of the control and the \lean model, with the cost. $\lambda=10$ for Qwen2-0.5B, 3 for the other two.}
\label{fig:llm}
\end{figure*}

\paragraph{Effect on language models}
This part is a preliminary extension, smaller in scale than the vision results: three models of 0.5 to 1.5 billion weights, one epoch on one million WikiText-2 tokens, perplexity as the metric. Weight sensitivity in transformers is skewed in the same way, and AttentionBreaker \cite{das2024attentionbreaker} and SilentStriker \cite{silentstriker2025} exploit it with a few dozen flips. Table~\ref{tab:llm} and Figure~\ref{fig:llm} apply the same objective to the linear layers of Qwen2-0.5B, Llama-3.2-1B and Qwen2-1.5B, fine-tuned for one epoch on one million WikiText-2 tokens against a control fine-tuned identically with $\lambda=0$; HIGH and MID are measured against the control's top-1\% and top-9\% thresholds, as for the CNNs. The HIGH set shrinks 1.7 to 5.3 times (5.3 is Llama at $\lambda=10$, at a 2.5 perplexity cost; 1.7 to 3.4 at the $\lambda$ we would deploy, Figure~\ref{fig:floor}c) and the MID set by about the same factor, and the cost is 0.6 perplexity at the $\lambda$ we would deploy ($\lambda=10$ on Qwen2, $\lambda=3$ on Llama; the penalty is summed over weights, so a larger model needs a smaller coefficient for the same pressure). Post-training INT4 improves on the two smaller models, from 21.1 to 19.3 on Qwen2-0.5B and from 21.4 to 18.9 on Llama, for the same reason as in Section~\ref{sec:results}: the weights the penalty removes from the tail are the ones that set the quantization grid. Qwen2-1.5B shows the other side of the knob: the penalty is summed over 1.3 billion weights, so $\lambda=3$ there is as strong as $\lambda=10$ on Llama, the tail shrinks 3.4 times but perplexity rises by 2.1 and INT4 does not improve; a smaller $\lambda$ is the right setting for a model of that size. Table~\ref{tab:llmflip} puts the three models under the same progressive sign-bit attack as the CNNs, with and without a 1\% code chosen by each model's own damage ranking. Without a code all three are destroyed within 10 flips. With the code the picture depends on size. On Llama-3.2-1B the \lean model is worse than the control for the first 20 flips (21 against 12.5 perplexity at 10 flips) and 3 times better from 50 flips on (42 against 122 at 50, 61 against 210 at 100), which is the regime a coded attacker has to work in. On the smaller Qwen2-0.5B a 1\% code is not enough for either model beyond 20 flips; \lean is at its clean perplexity after 10 flips where the control has already doubled, and still usable at 20 where the control is gone. On Qwen2-1.5B, where $\lambda=3$ was already too strong, the attack shows the same thing: both models hold 20 flips with the code and both are gone at 50, and the \lean model is slightly worse at 20 (16.3 against 11.2), so an over-strong $\lambda$ buys nothing. Where $\lambda$ is set so that perplexity moves by less than one point, the gain is the same kind as on the CNNs, the code removes the weights the attacker wants and $\lambda$ thins what is left.

\paragraph{Limitations}
\lean needs training data and about 1.5 times the cost of a plain fine-tune, and the low-bit format needs a second one. Our adaptive attacker is white-box and knows the code, but it is greedy; a search that plans across codewords could do better, and the protection results rest on the interleaving assumption of Section~\ref{sec:background}. Most entries are single runs; the storage comparison and the ResNet-18 CIFAR-10 reference model have three seeds (the experimental setup of Section~\ref{sec:results}), and the accuracy differences of 0.1 to 0.5 points in Table~\ref{tab:sens} should be read with that in mind; the 500-flip outcome of the coded ResNet-18 sits at a threshold and varies with the training seed, while the 300-flip outcome does not. The evaluation is of adaptive flips; random faults are measured at one bit error rate only (Table~\ref{tab:ecc}). The robust-training baselines of Table~\ref{tab:def} were fine-tuned with our ten-epoch schedule rather than their original recipes. The tail term is not a defense on its own: Table~\ref{tab:ablate} shows that alone it lowers the gradient without lowering the damage, which is why \lean always pairs it with the range term, and why the real-flip check should accompany any proxy of this kind. The ImageNet models cost more bits in the low-bit security mode: ResNet-50 needs the 3\% tier and the $t=79$ code, 4.09 bits per weight, to hold 500 flips, and ResNet-18 at that cost is still 14 points below clean. A $\lambda$ fine-tune that targets the deployed format more directly, or a wider tier chosen during training rather than at packing, are the natural next steps. In this paper $\lambda$ is the only knob, and it moves the whole tail at once. The objective has other parameters that could shape the distribution more selectively and that we have not explored: the tail fraction that defines the set $T$ (0.5 to 2\% gives the same HIGH bin, but a per-bin target could act on MID and HIGH separately), the relative weight $\beta$ of the range and tail terms, the clipping quantile $q_{99}$, and the learning rate and number of epochs, which Table~\ref{tab:inset} already shows to set how far the tail moves for a fixed $\lambda$. A family of such controls, each tied to one part of the distribution, is the natural extension of the one-knob procedure presented here.

\section{Related Work}
\label{sec:related}
\lean sits between three lines of work, and takes a different position from each.

\paragraph{Robust training against bit flips}
Piecewise clustering \cite{he2020pc} and quantization-range learning (WCAT, \cite{chitsaz2023wcat}) make the stored values of a network tolerant to flips by constraining weight magnitudes; Forget-and-Rewire \cite{nazari2024far} reroutes the most important connections so that single flips matter less; Hessian-aware training \cite{pranto2025hessian} regularizes curvature. All of these change the network so that a flip hurts less, and all of them, like \lean, are at chance within 50 flips of an adaptive attacker when no code is present (Table~\ref{tab:def}). WCAT is the closest to the range term of \lean, since both bound the per-layer quantization range; Table~\ref{tab:def} separates them under the same code (WCAT, one run, 75\% at 500 flips; \lean 94\% in two of three seeds), and the real-flip rows of Table~\ref{tab:ablate} show what the tail term adds on top of range control. \lean differs from all four in what it optimizes: not robustness to flips in general but the size of the set that a selective code has to cover, measured in the storage format.

\paragraph{Gradient-norm regularization}
$P_{\mathrm{tail}}$ belongs to an old family: double backpropagation \cite{drucker1992double} penalizes the input gradient, gradient-norm regularization and sharpness-aware minimization \cite{foret2021sam} penalize the weight gradient or the loss in a neighbourhood. \lean differs in two details that the ablation shows are the ones that matter for stored weights: the penalty is top-$k$, so only the tail is pushed, and it is weighted by the storage-format step $\Delta_\ell$, which ties it to the range term. A plain gradient-norm fine-tune under the same budget is the natural further baseline, and Tables~\ref{tab:ablate} and \ref{tab:def} include it: at the largest coefficient the accuracy tolerates it leaves the tail at 0.64\% and the coded model breaks at 200 flips like the normal one, so the tail-selective, storage-weighted form is not a generic gradient smoother.

\paragraph{Selective and in-place protection}
Selective and embedded codes such as in-place zero-space protection \cite{guan2019inplace} and E3-CODE \cite{kamineni2026e3} reduce the parity cost of error correction by protecting only the weights that matter or by hiding the parity in the weights themselves. Our contribution is upstream of them: it makes the set they protect smaller, and is compatible with any of them. Detection schemes such as RADAR \cite{li2020radar} and Aegis \cite{wang2023aegis} catch flips at run time instead of correcting them; a correcting code was chosen here because it needs no response path, but the smaller sensitive set would lower the cost of a checksum in the same way.

\paragraph{Low-bit quantization}
LSQ \cite{esser2020lsq} learns the quantizer step; KURE \cite{shkolnik2020kure} shapes the weight distribution for robustness across bit widths; HAWQ \cite{dong2019hawq} assigns bit widths per layer from Hessian information; SpQR \cite{dettmers2023spqr} and SqueezeLLM \cite{kim2023squeezellm} keep a small set of outlier or sensitive weights at high precision inside a low-bit model, which is per-weight precision of the same kind as our tiers. Our tiered format assigns bit widths per weight by stored-format sensitivity, and its difference from those methods is upstream: $\lambda$ makes the high-precision set smaller before the format is chosen, and the per-channel INT4 we use as the LLM baseline is a deliberately simple one (GPTQ \cite{frantar2023gptq} and AWQ \cite{lin2024awq} would lower both rows of Table~\ref{tab:llm}). At a uniform bit width \lean matches LSQ rather than beating it (Table~\ref{tab:storage}); the gain comes from the ordering, which the shrunken sensitive set makes affordable. Generating weights from a compact seed \cite{rahaman2025secure} is an orthogonal route to storage efficiency that a reshaped sensitivity distribution does not preclude.

\paragraph{Attacks on large language models}
AttentionBreaker \cite{das2024attentionbreaker} and SilentStriker \cite{silentstriker2025} carry gradient-guided bit flips to transformers, where a few dozen flips suffice. The sensitivity structure they exploit is the one \lean shapes, and Section~\ref{sec:app} reports a first adaptation.

\section{Conclusion}
\label{sec:concl}
The weights of a trained neural network model are not equally sensitive with respect to its functional behavior. The shape of their sensitivity distribution sets what a bit-flip attacker needs, what a code must cover, and how many bits each weight has to keep. Existing methods do not optimize the shape of the weight sensitivity distribution. We have presented \lean that allows us to modulate the weight sensitivity distribution, leading to varying numbers of weights in different weight sensitivity bins. For example, it helps greatly reduce the number of weights in the highest sensitivity bins by moving them to lower sensitivity bins. Such a shift in weight sensitivity distribution may result in lower cost for protection against bit-flip attacks as well as increased efficiency in model weight storage (e.g., sensitivity-guided variable precision quantization of weights) and energy/performance for computing activation at each node.  

One composite regularizer on the two factors of the damage, the flip step and the gradient, controlled by a single coefficient, moves the sensitive tail 5 to 13 times lower on nine vision models for an accuracy change under 0.5 points and 1.7 to 5.3 times on three language models, and the amount follows $\lambda$ monotonically. The term works as a ten-epoch fine-tune or in the last epochs of training, and it stops at a floor set by the model itself, its per-layer outlier ratios and the curvature around its early layers. The reshaped distribution then does the rest. A 1\% code holds ResNet-18 through 300 to 500 adaptive flips where three of four published defenses fail by 300 and WCAT falls to 75\% at 500, the parity for a fixed guarantee drops 3.7 times, a 3.29-bit sensitivity-ordered format, which the small tail makes affordable, stores the ImageNet ResNet-50 weight payload 2.4 times smaller than INT8 above the accuracy of LSQ, and a small interleaved code lets the low-bit models survive too. The targeted attack on the ImageNet ResNets, which moves through weights the mixed-data sensitivity does not rank, is the case the current objective does not reach, and shaping the distribution under a class-aware sensitivity is the next step, as is control through parameters other than $\lambda$, such as a per-bin target or the balance of the two terms, so that different parts of the distribution can be moved independently.

\bibliographystyle{IEEEtran}
\bibliography{lean_arxiv}
\end{document}